\documentclass{article} % For LaTeX2e
\usepackage[preprint]{rlc}

\usepackage{amssymb}            % Defines common symbols like \mathbb R
\usepackage{mathtools}          % Extends amsmath, providing common math tools
\usepackage{mathrsfs}           % Enables \mathscr, which can work in cases that \mathcal does not
\usepackage{graphicx}           % For including images
\usepackage{subcaption}         % Allows for the use of subfigures and subcaptions
\usepackage[space]{grffile}     % For spaces in image names
\usepackage{url}                % For displaying URLs
\usepackage{float}
\usepackage{tikz}
\usepackage{todonotes}
\usepackage{svg}
\usepackage{multirow}

\title{Contrastive Sparse Autoencoders for Interpreting Planning of Chess-Playing Agents}

\author{Yoann Poupart  \\
    yoann.poupart@ens-lyon.org \\
    ENS de Lyon
}

\begin{document}

\maketitle

\begin{abstract}
AI led chess systems to a superhuman level, yet these systems heavily rely on black-box algorithms. This is unsustainable in ensuring transparency to the end-user, particularly when these systems are responsible for sensitive decision-making. Recent interpretability work has shown that the inner representations of Deep Neural Networks (DNNs) were fathomable and contained human-understandable concepts. Yet, these methods are seldom contextualised and are often based on a single hidden state, which makes them unable to interpret multi-step reasoning, e.g. planning. In this respect, we propose contrastive sparse autoencoders (CSAE), a novel framework for studying pairs of game trajectories. Using CSAE, we are able to extract and interpret concepts that are meaningful to the chess-agent plans. We primarily focused on a qualitative analysis of the CSAE features before proposing an automated feature taxonomy. Furthermore, to evaluate the quality of our trained CSAE, we devise sanity checks to wave spurious correlations in our results.
\end{abstract}

\section{Introduction}

Chess is one of the very first domains where superhuman AI shined, first with DeepBlue \citep{campbell2002deep} and more recently with Stockfish \citep{nasu2018nnue} and AlphaZero \citep{silver2018general}. While the design of these superhuman programs is intended to gain performances, e.g. by optimising the tree search, the node evaluation or the training procedure, a lot remains to be done to understand the intrinsic processes that led to these performances truly. In this respect, the first component to decipher is thus the DNN heuristic that guides the tree search. While DNNs are often thought of as black-box systems, they learn a basic linear representation of features. During the last decade, arguments to support this hypothesis have been demonstrated repeatedly for language models \citep{Mikolov2013LinguisticRI, Burns2022DiscoveringLK,Tigges2023LinearRO} but also vision models \citep{Radford2015UnsupervisedRL,Kim2017InterpretabilityBF,Trager2023LinearSO} and others \citep{Nanda2023EmergentLR,Rajendran2024LearningIC}. This strong hypothesis also transferred to chess \citep{McGrath2022Acquisition}, showing that traditional concepts like "attacks" or "material advantage" were linearly represented in the latent representation of the model.

 In this work, we focus on the open-source version of Alpha Zero, Leela Chess Zero \citep{pascutto2019leela}, interpreting the neural network heuristic in combination with the tree search algorithm. In particular, we extend the dynamic concepts introduced in \cite{schut2023bridging}.
We state our contributions as follows:
\begin{itemize}
    \item New dictionary architecture to encourage the discovery of differentiating features between latent representations
    \item Automated sanity checks to ensure the relevance of our dictionaries
    \item Discovery and interpretation of new strategic concepts creating a feature taxonomy
\end{itemize}

\begin{figure}[H]
    \centering
    \begin{tikzpicture}
        \draw (0, 0) node[inner sep=0] 
        {\includegraphics[width=0.5\textwidth]{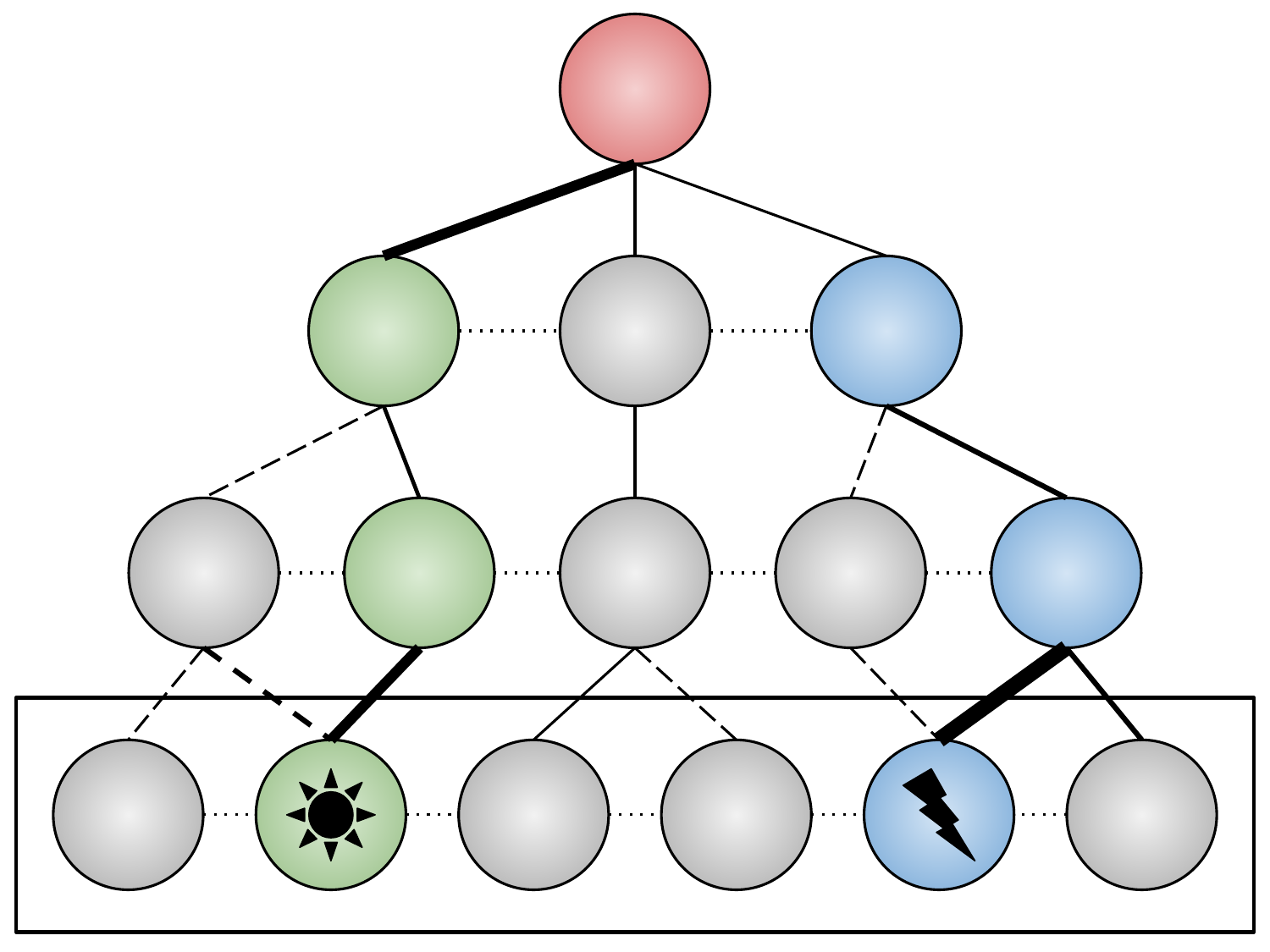}};
        \draw (0, 3) node {$s_0$};
        \draw (1.8, 1.7) node {$\mathbb{S}^-_{\leq 3}(s_0)$};
        \draw (-1.8, 1.7) node {$\mathbb{S}^+_{\leq 3}(s_0)$};
    \end{tikzpicture}
    
    \caption{Better viewed in colour. Our proposed framework aims to retrieve planning concepts, represented as icons at the bottom. For that, we analyse the 
    plans of a chess-playing agent. A sampling of an optimal trajectory $\mathbb{S}^-_{\leq 3}(s_0)$ (in green) and a suboptimal trajectory $\mathbb{S}^+_{\leq 3}(s_0)$ (in blue) from a root node $s_0$. The star represents a concept meaningfully to the optimal trajectory while the lightning represents a concept relevant to the suboptimal trajectory.  }
    \label{fig:dynamical_concepts}
\end{figure}
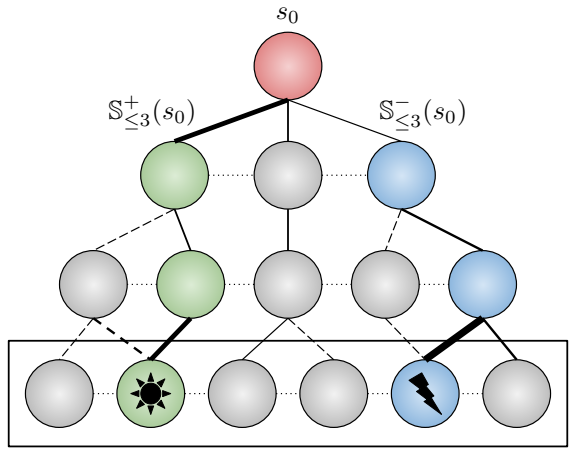

Figure \ref{fig:dynamical_concepts} summarises our approach and illustrates our aim at disentangling planning concepts. With this paper, we release the code\footnote{
Released at \href{https://github.com/Xmaster6y/lczero-planning}{https://github.com/Xmaster6y/lczero-planning}.
} used to create the datasets and to discover and analyse concepts.

\section{Background}
\label{sec:background}
\subsection{Chess Modelling}

\begin{figure}[H]
     \centering
     \begin{subfigure}[b]{0.26\textwidth}
         \centering
         \includegraphics[width=\textwidth]{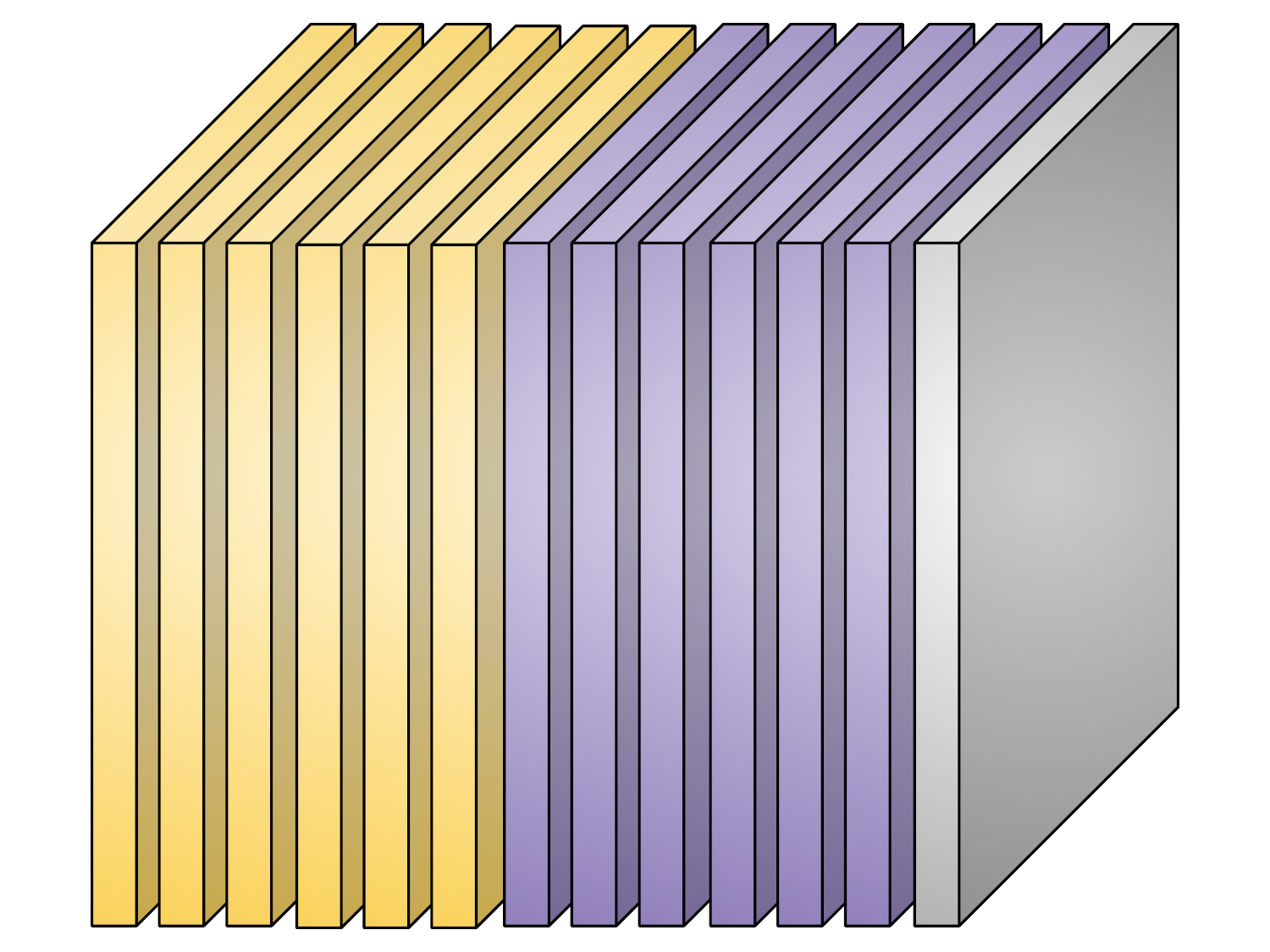}
         \caption{Board encoding}
         \label{fig:background_a}
     \end{subfigure}
     \hspace{8em}
     \begin{subfigure}[b]{0.26\textwidth}
         \centering
         \includegraphics[width=\textwidth]{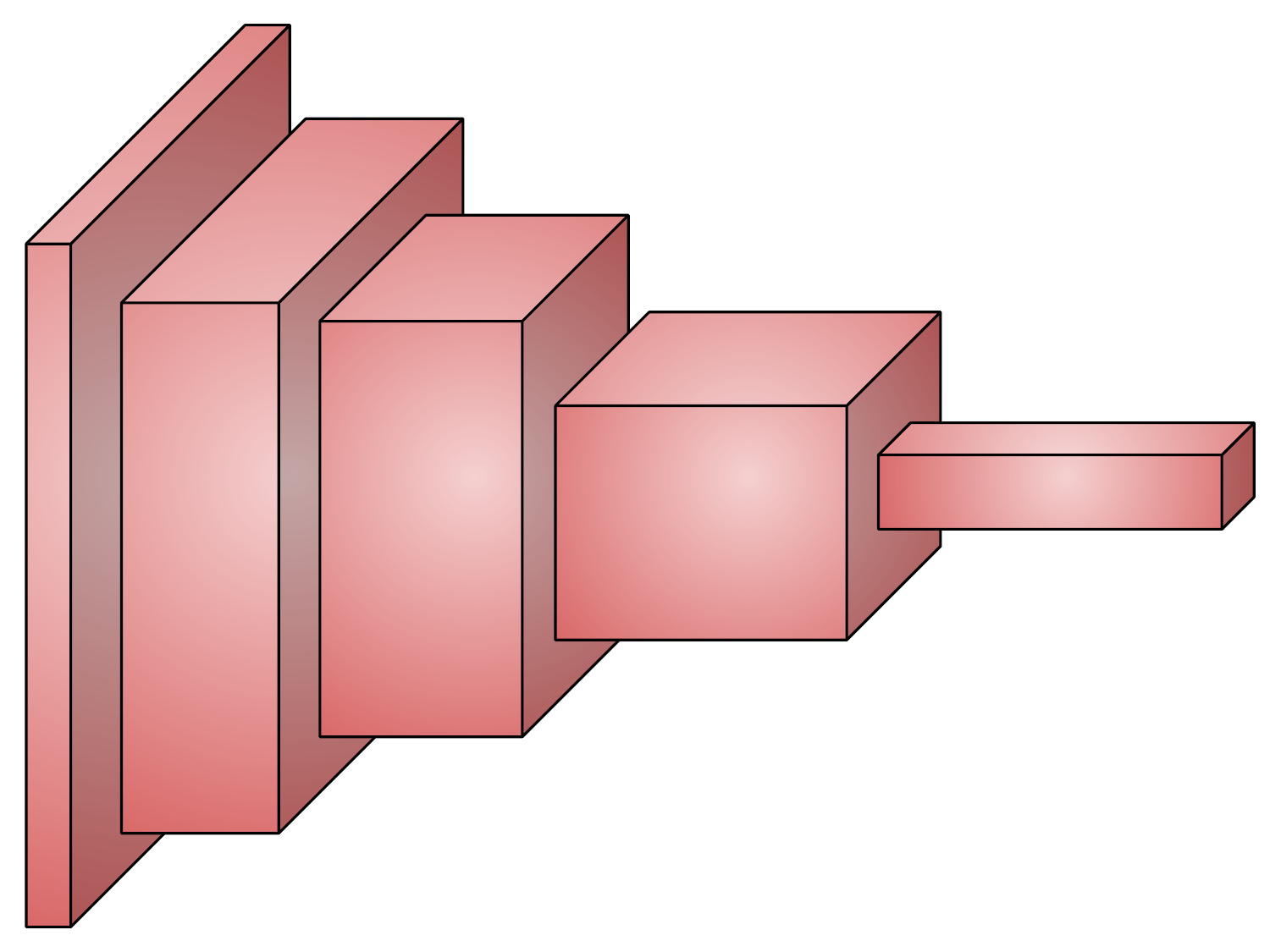}
         \caption{Network backbone}
     \end{subfigure}
     \\
     \begin{subfigure}[b]{0.26\textwidth}
         \centering
         \begin{tikzpicture}
            \draw (0, 0) node[inner sep=0] 
            {\includegraphics[width=\textwidth]{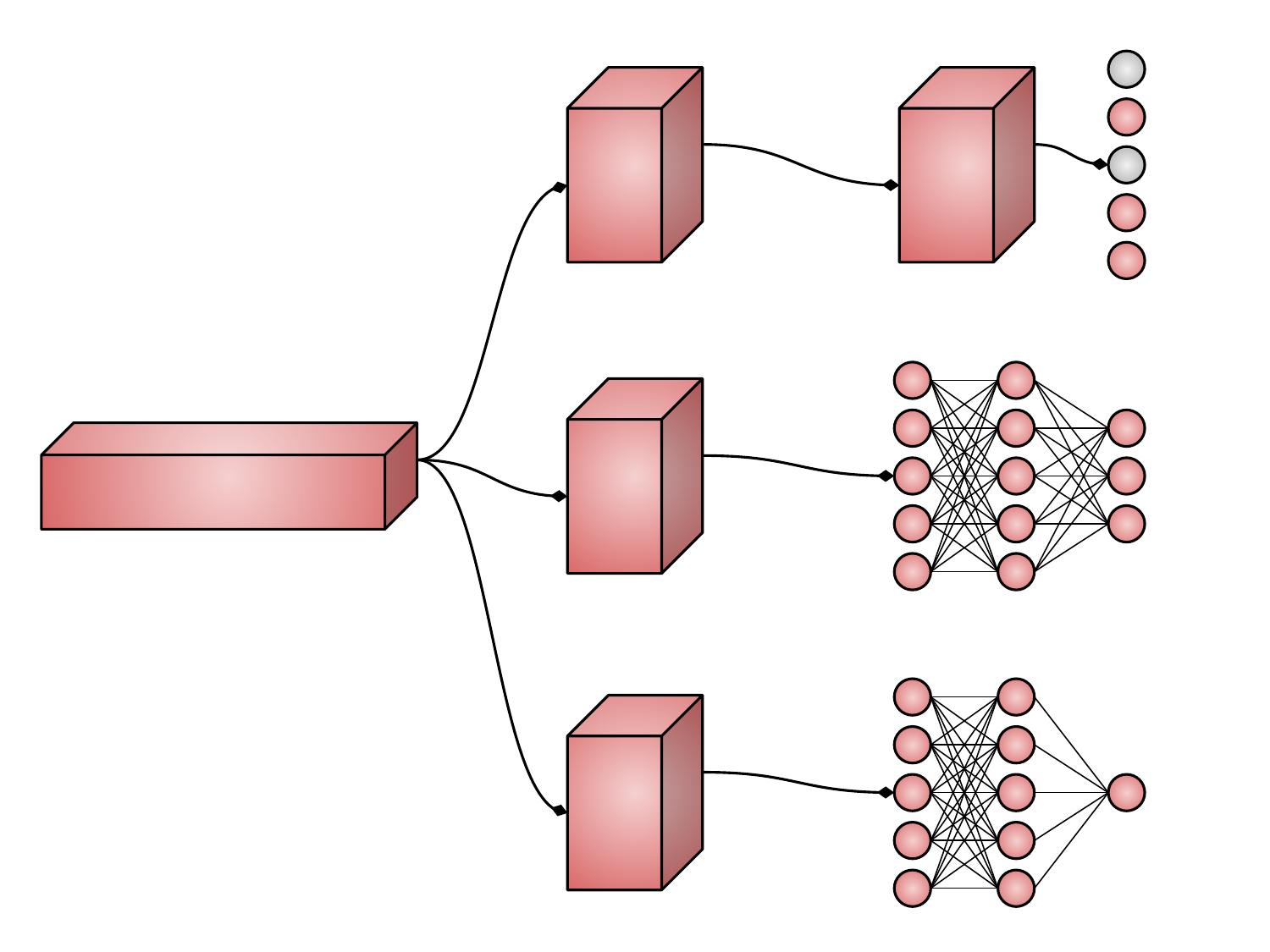}};
            \draw (2.3, 1) node {$\mathcal{P}_\theta(s)$};
            \draw (2.3, 0) node {$\mathcal{W}_\theta(s)$};
            \draw (2.3, -1) node {$\mathcal{M}_\theta(s)$};
        \end{tikzpicture}
         \caption{Heads prediction}
     \end{subfigure}
     \hspace{8em}
     \begin{subfigure}[b]{0.26\textwidth}
         \centering
         \includegraphics[width=\textwidth]{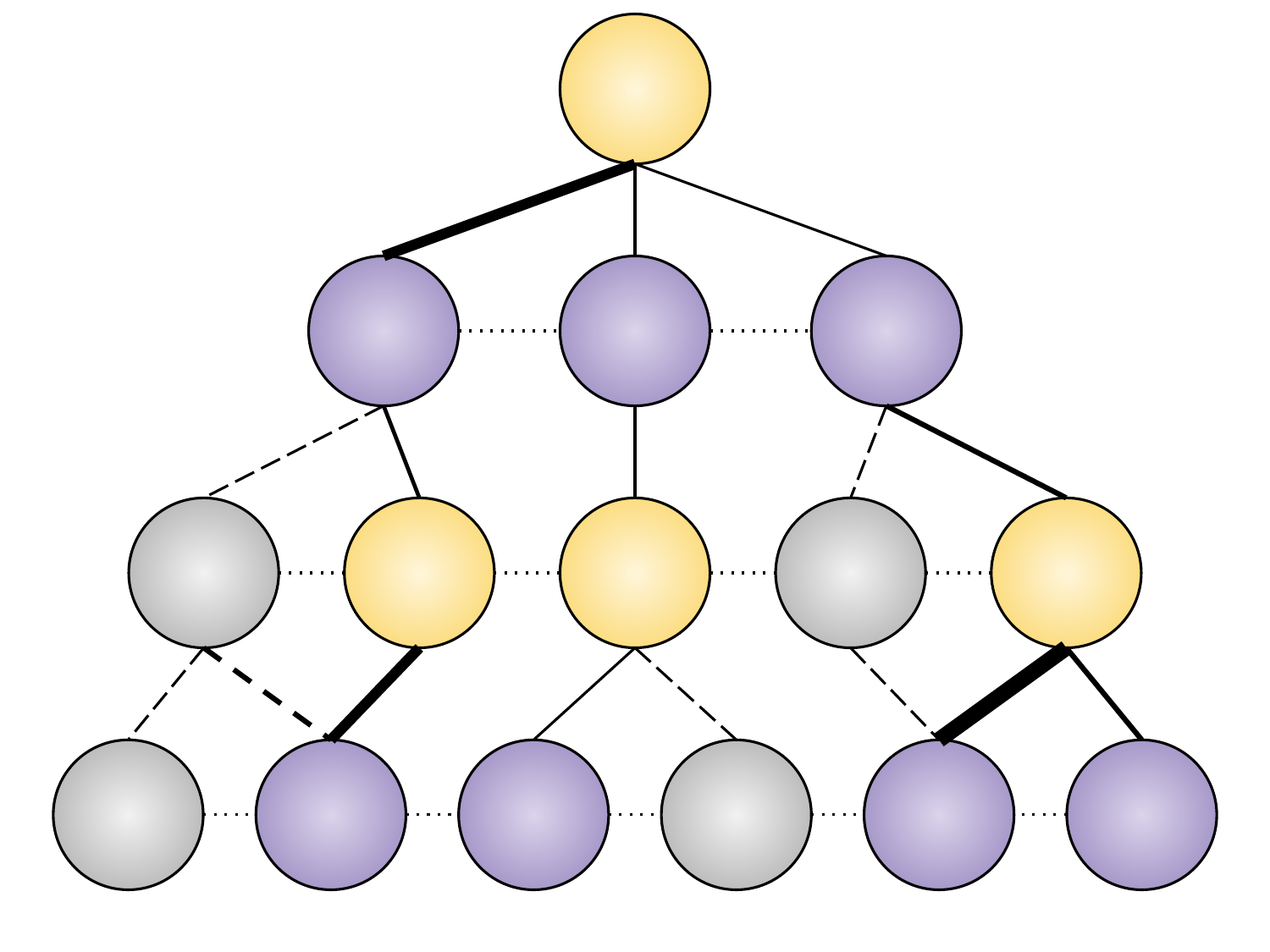}
         \caption{MCTS}
     \end{subfigure}
     \caption{Modelling components; first, the boards are encoded into planes (a) and fed to the network backbone (b).
     The different heads use the extracted features to make heuristic predictions (c) guiding the MCTS when encountering new nodes (d).}
     \label{fig:background}
\end{figure}
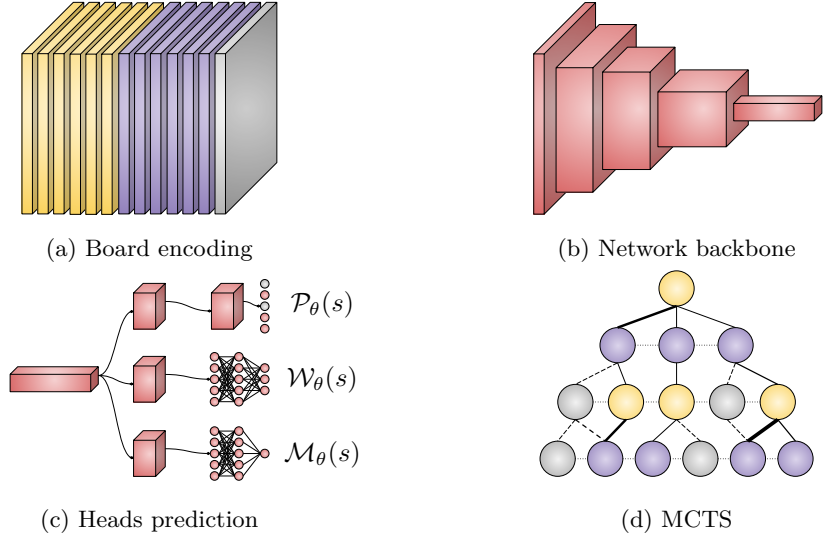

\paragraph{Heuristic network} The studied agent, introduced as AlphaZero \citep{silver2018general}, is a heuristic network used in a Monte-Carlo tree search (MCTS) \citep{Coulom2006EfficientSA, kocsis2006bandit}. The network is traditionally trained on self-play to collect data, i.e. the network is frozen and plays against a duplicate version of itself. After the collection phase, the network is trained to predict a policy vector for the next move based on the MCTS statistics and a current state value based on the outcomes of the played games. Here, more specifically, the full network $\mathcal{F}_\theta$, parametrized by $\theta$, can be describe as a tuple,
\begin{equation}
    \mathcal{F}_\theta(s) = \left[\mathcal{P}_\theta(s),\,\mathcal{W}_\theta(s),\, \mathcal{M}_\theta(s)\right],
\end{equation}
with $\mathcal{P}_\theta(s)$ the policy vector, $\mathcal{W}_\theta(s)$ the win-draw-lose probability and $\mathcal{M}_\theta(s)$ the moves left.
The three heads share a Squeeze-and-Excitation (SE) backbone \citep{hu2019squeezeandexcitation}, based on ResNet \citep{he2016deep}.
The state $s$ fed to the network is made of the current board as well as the 7 previous boards.
These boards are decomposed into one-hot planes that we describe in the next paragraphs.
The computation process is illustrated in figure \ref{fig:background}; for more details, we refer the reader to the exact implementation in \citep{pascutto2019leela}.

\paragraph{Tree-search} The AlphaZero \citep{silver2018general} and its open-source version LeelaZero \citep{pascutto2019leela} are based on evaluation and tree search similar to Stockfish NNUE. The search algorithm is based on MCTS \citep{Coulom2006EfficientSA, kocsis2006bandit} using a slightly modified version of the upper bound confidence of the PUCT algorithm \citep{Rosin2011MultiarmedBW}, equation \ref{eq:upper_confidence_boundary}.

\begin{equation}
\label{eq:upper_confidence_boundary}
    U(s,a)=Q(s,a)+c_{\rm puct}\cdot P(s,a) \cdot \dfrac{\sqrt{\sum_b N(s,b)}}{1+N(s,a)}
\end{equation}

Here, we focused on the policy $P(s,a)=\mathcal{P}_\theta(s,a)$ directly outputted by the network. We further detail the computation of the $Q$-values and their links to the WDL head $W_\theta(s,a)$ and the ML head $M_\theta(s,a)$ in the appendix \ref{sec:addi_chess_modelling}.

\subsection{Discovering Concepts}

\paragraph{Sparse autoencoders}
While linear probing \citep{alain2018understanding} requires labelled concepts, sparse autoencoders are an efficient tool for discovering concepts at scale without supervision, which were introduced concurrently in \cite{Cunningham2023SparseAF} and \cite{bricken2023monosemanticity}.
The fundamental idea is to decompose the latent activations $h$ on a minimal set of features, formulated as the minimisation of
\begin{equation}
    ||h - Df||_2^2+\lambda||f||_0.
\end{equation}
$D$ is the feature dictionary and $f$ is the feature decomposition with $f\geq0$ for the combination view. In practice, sparse autoencoders (SAEs) have been proposed to solve sparse dictionary learning and have already proven to find a wide range of interpretable features \citep{bricken2023monosemanticity}.
In their simplest form, with only one hidden layer, the architecture can be described as
\begin{align}
    f &= \text{ReLU}(W_{\rm e}h+ b_{\rm e}), \
    \hat{h} &= W_{\rm d}f + b_{\rm d}.
\end{align}
Where the encoder weights ($W_{\rm e}$, $b_{\rm e}$) and decoder weights ($W_{\rm d}$, $b_{\rm d}$) are trained using an MSE reconstruction loss with $l_1$ penalisation to incentivize sparsity:

\begin{align}
    \label{eq:sae_trad_loss}
    \mathcal{L}_{\rm SAE}=\mathbb{E}_h\left[||h-\hat{h}||_2^2 +
    \lambda ||f||_1
    \right]
\end{align}

We describe in appendix \ref{sec:training_details} some additional architectural changes and hyperparameters we used and how we evaluated those.

\paragraph{Dynamical concepts}
 While traditional concepts only rely on a single position \citep{McGrath2022Acquisition}, dynamical concepts consider sequences of states and are still discoverable using linear probing \citep{schut2023bridging}. In order to find these concepts, we need to
 consider an optimal rollout, according to the chosen sampling method, $\mathbb{S}^+_{\leq T}(s_0)~=~(s^+_1, s^+_2, ..., s^+_T)$ with $T$ being the maximal depth considered starting at state $s_0$.
This rollout is associated with other sub-optimal rollouts $\mathbb{S}^-_{\leq T}~=~(s^-_1, s^-_2, ..., s^-_T)$. A linear probe can then be trained to differentiate the origin set of a state $s$ using the model's hidden state $h$; the process is illustrated in Figure \ref{fig:dynamical_concepts}.
% \begin{figure}[H]
%     \centering
%     \begin{tikzpicture}
%         \draw (0, 0) node[inner sep=0] 
%         {\includegraphics[width=0.5\textwidth]{figures/Dynamical Concepts.pdf}};
%         \draw (0, 3) node {$s_0$};
%         \draw (1.8, 1.7) node {$\mathbb{S}^-_{\leq 3}(s_0)$};
%         \draw (-1.8, 1.7) node {$\mathbb{S}^+_{\leq 3}(s_0)$};
%     \end{tikzpicture}
    
%     \caption{Better viewed in colour. A sampling of an optimal trajectory $\mathbb{S}^-_{\leq 3}(s_0)$ (in green) and a suboptimal trajectory $\mathbb{S}^+_{\leq 3}(s_0)$ (in blue) from a root node $s_0$. Dynamical concepts (represented with symbols) abstract the differentiating features between the two trajectories and are supposedly responsible for the models' choices, i.e. planning.}
%     \label{fig:dynamical_concepts}
% \end{figure}

\section{Methods}

\subsection{Disantangling Planning Concepts}
\label{sec:method_dis}

\begin{figure}[H]
     \centering
     \begin{subfigure}[b]{0.49\textwidth}
         \centering
         \begin{tikzpicture}
            \draw (0, 0) node[inner sep=0] 
            {\includegraphics[width=\textwidth]{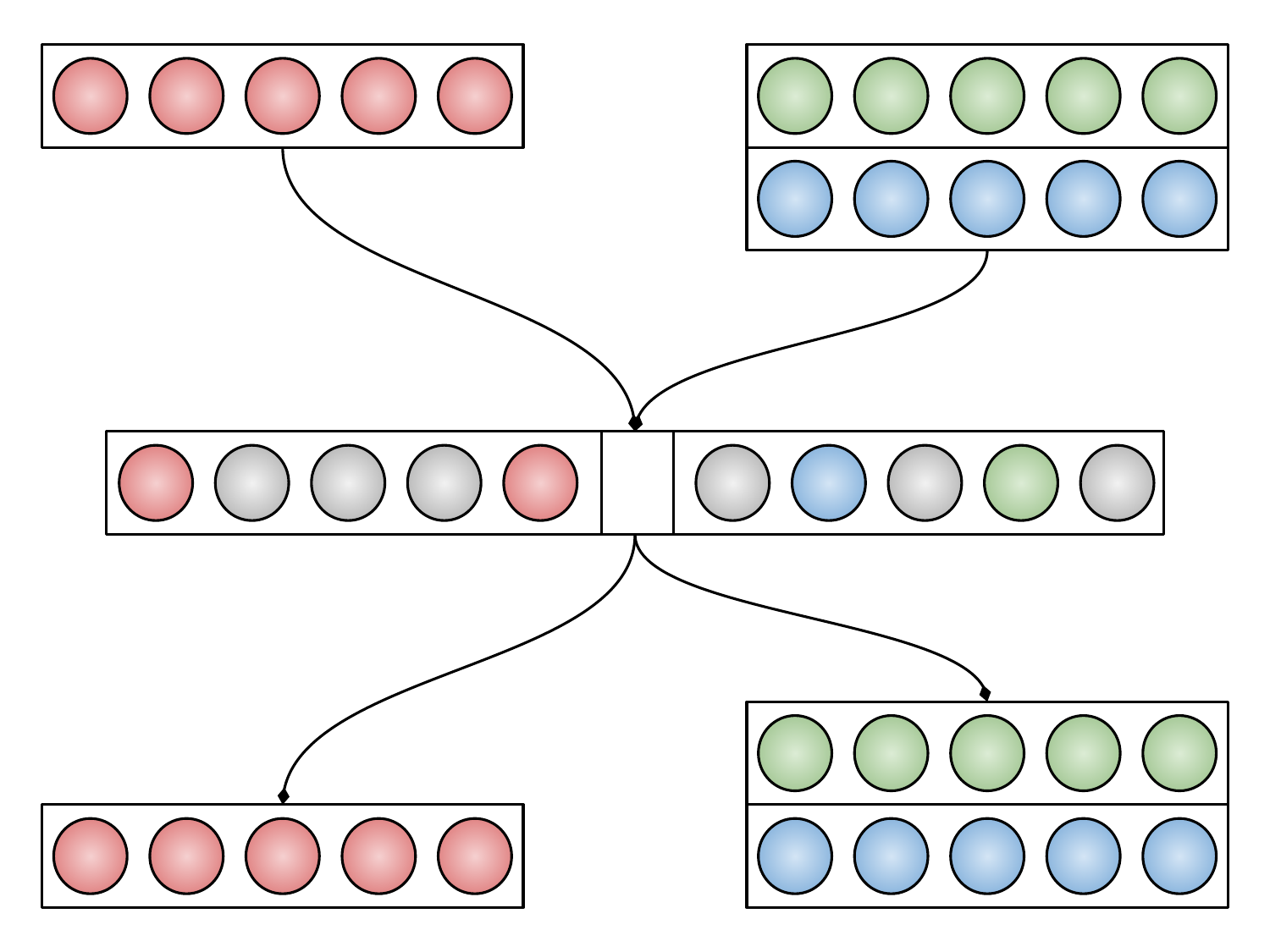}};
            \draw (-4, 2.3) node {$h(s_0)$};
            \draw (0.1, 2.3) node {$h(s_t^+)$};
            \draw (0.1, 1.7) node {$h(s_t^-)$};
            \draw (-3.3, 0) node {$c$};
            \draw (3.3, 0) node {$d$};
            \draw (-4, -2.2) node {$\hat{h}(s_0)$};
            \draw (0.1, -2.2) node {$\hat{h}(s_t^-)$};
            \draw (0.1, -1.6) node {$\hat{h}(s_t^+)$};
        \end{tikzpicture}
         \caption{Contrastive SAE}
         \label{fig:sae_contrast}
     \end{subfigure}
        \hfill
     \begin{subfigure}[b]{0.49\textwidth}
         \centering
         \begin{tikzpicture}
            \draw (0, 0) node[inner sep=0] 
            {\includegraphics[width=\textwidth]{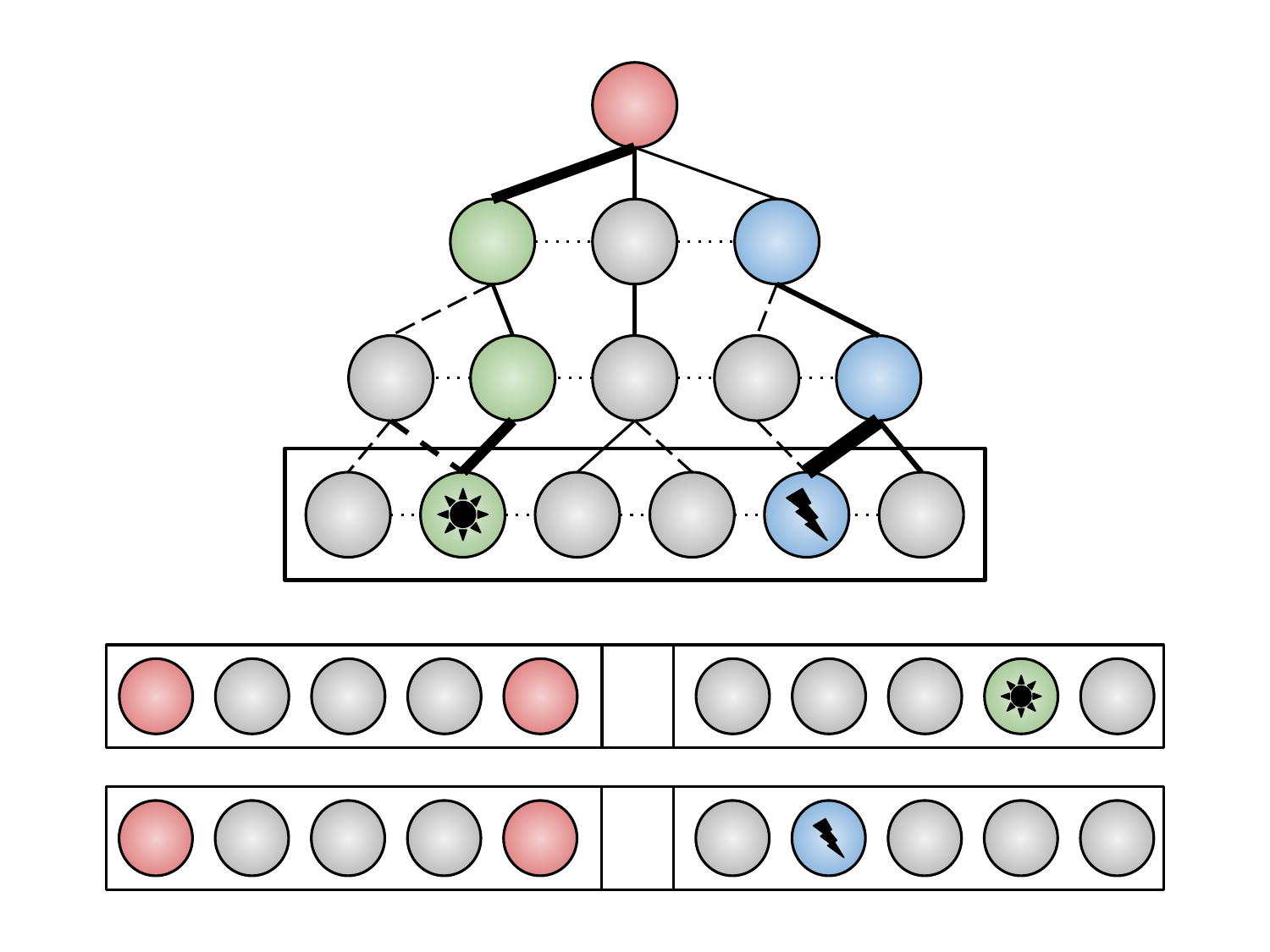}};
            \draw (0, 2.7) node {$s_0$};
            \draw (1.2, 2) node {$\mathbb{S}^-_{\leq 3}(s_0)$};
            \draw (-1.2, 2) node {$\mathbb{S}^+_{\leq 3}(s_0)$};
            \draw (3.5, -1.2) node {$d^+$};
            \draw (3.5, -2.1) node {$d^-$};
        \end{tikzpicture}
         \caption{Rollouts concepts extraction}
         \label{fig:sae_contrast_dyn}
     \end{subfigure}
     \caption{Better viewed in colour. (a) Contrastive SAEs are trained using a contrast of an optimal trajectory (green) and suboptimal trajectories (blue). They take in input the root hidden state $h(s_0)$ and a subsequent node's hidden state $h(s_t^\pm)$. The $c$-features are represented in red, and the $d$-features are in blue and green. (b) Schematic view of concepts extraction from different rollouts. The dynamical concepts from the rollout $\mathbb{S}^+_{\leq 3}(s_0)$ is extracted in $d^+$ and for $\mathbb{S}^-_{\leq 3}(s_0)$ in $d^-$.}
     \label{fig:comp_contrast}
\end{figure}
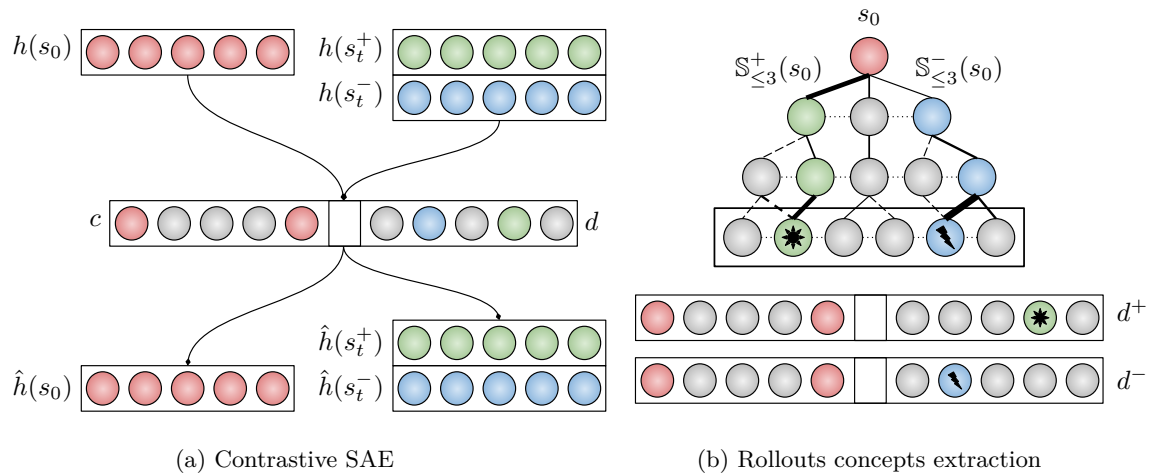

The basic idea proposed here is to study a latent space vector in contrast with others. The intuition is that we want to know what additional concepts are present in subsequent states. So, for a depth $t$, we use a pair of vectors defined as a concatenation of the search root $s_0$ with $s_t^+$ from the optimal rollout and $s_t^-$ from a suboptimal rollout; similarly to \cite{schut2023bridging}.
\begin{align}
    h^+&=[h(s_0);h(s_t^+)]\\
    h^-&=[h(s_0);h(s_t^-)]
\end{align}
We introduce a feature constraint in order to train SAEs with a contrastive loss, equation \ref{eq:sae_contrast_loss}. By dividing the feature dictionary into a set of common features $c$ and a set of differentiating features $d$, we can separate the $s_0$ dependence and focus on planning concepts contained in $d$. In practice, the separation is made using tensor concatenation $f=[c;d]$ as illustrated in the figure \ref{fig:sae_contrast}.

\begin{equation}
    \label{eq:sae_contrast_loss}
    \mathcal{L}_{\rm contrast}=\mathbb{E}_h
    \left[||c^+-c^-||_1+||d^+\odot d^-||_1\right]
\end{equation}

In order to concentrate the $s_0$ dependence into the $c$-features, we added an additional SAE loss term (reconstruction and sparsity) to reconstruct $h(s_0)$ from $c^+$ and $c^-$.
Additionally, to ensure that the $d$-features account for differentiability, we train a linear probe on this intermediate representation of our SAEs using the binary cross-entropy, equation \ref{eq:clf_pm_loss}. We present the results as part of our first sanity checks in the section \ref{sec:sanity_checks}.

\begin{equation}
    \label{eq:clf_pm_loss}
    \mathcal{L}_{\pm}=\mathbb{E}_h
    \left[
    -\log\left\{\mathcal{P}(d^+)\right\}
    -\log\left\{1-\mathcal{P}(d^-)\right\}\right]
\end{equation}

\subsection{Concepts Interpretation}
\label{sec:method_int}

\begin{figure}[H]
     \centering
     \includegraphics[width=0.45\textwidth]{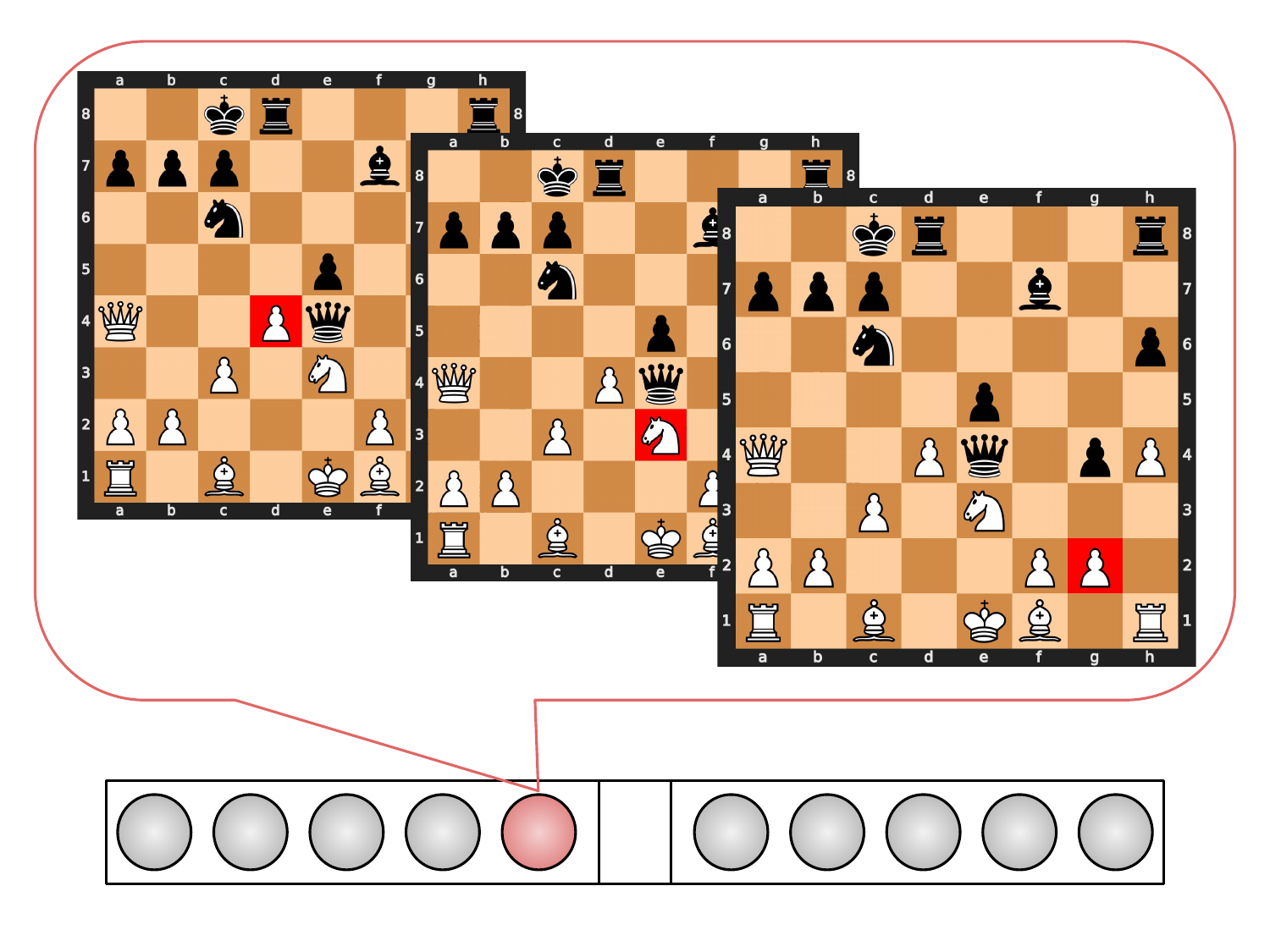}
     \caption{(a) Illustration of the process of interpreting a feature using activation maximisation. The most activated samples are retrieved and analysed. (b) In order to compare a pair of features, the first indicator is the correlation of the feature activation (right). It is also possible to count common samples retrieved using activation maximisation.}
     \label{fig:data_act_max}
\end{figure}

\paragraph{Interpreting individual features} In order to decipher the nature of the learned dictionary features, a first qualitative analysis can be run using activation maximisation based on data sample \citep{chen2020concept}. As illustrated in figure \ref{fig:data_act_max}, for a given feature, it is possible to investigate the most activated samples. Here, the samples are latent pixels and thus can be visualised on the corresponding chess boards. It is thus possible to create a basic feature categorisation based on the samples they activate in and whether they activate on a wide or restricted range of samples. 

% \paragraph{Comparing features by pair}
% It is important to investigate the correlation between features, which is a simple proxy to understand basic interactions between features. This analysis can be run for the $c$-features and the $d$-features, which is illustrated in figure \ref{fig:act_max_comp}. We first present a sanity check on the $c$-features in section \ref{sec:sanity_checks} and expand $d$-features categorisation in \ref{sec:dyn_con_clustering}.
% This method is especially relevant when dealing with different latent spaces, e.g. from different models or layers.  In appendix \ref{sec:accross_models_layers}, we present a small investigation of the correlation between features from different layers and at different training stages.

\paragraph{Categorising concepts} While the learned features appear to be relatively interpretable, it does not scale well with respect to the required human labour. Recent work proposed automated methods to interpret models based on causal analysis \citep{Conmy2023TowardsAC},  using a language model interpreter \citep{bills2023language} or a multimodal model \citep{Shaham2024AMA}. Yet these methods are hard to supervise humanly and are adding an additional black box layer. We investigate a frugal alternative, creating an automated taxonomy of features using hierarchical clustering. To test this taxonomy, presented in section \ref{sec:dyn_con_clustering}, we propose a last sanity check based on the $c$-features in section \ref{sec:sanity_checks}.

\section{Experiments}
\label{sec:experiments}

\subsection{Sanity Checks}
\label{sec:sanity_checks}
% We justify our architecture choice by a will to separate dynamical concepts from root-related concepts. It is thus important to explore whether this proves true in practice. In this respect, we designed sanity checks to alleviate trivial errors. Furthermore, we discuss the choice of hyperparameters and trade-offs and report key metrics in the appendix \ref{sec:training_details}.

\paragraph{Partitioned features}

Then, to understand the coarse-grained difference between $c$-features and $d$-features, we compute a set of metrics reported in the table \ref{tab:sanity_metrics}. The metrics are computed on unseen examples (\texttt{test}) similarly to \texttt{validation} but were not optimised against. 
We reported additional metrics in the appendix \ref{sec:training_details}.

\begin{table}[H]
    \centering
    \begin{tabular}{l|c|c|c|c|c|c|c|}
    Metric &  $F<0.1\%$ &  $F>10\%$ &$H(A_s)$ &$H(A_t)$ & $F_1(\mathcal{P})$ & $P(\mathcal{P})$ & $R(\mathcal{P})$ 
    \\
    \hline
    $c$-features &153&58&
    2.18&2.81&
    0.537&0.541&0.534\\
    $d$-features &0&119&
    2.33&3.24&
    0.566&0.575&0.557\\
    $f$&153&177&
    2.25&3.02&
    0.578&0.584&0.571
    \end{tabular}
    \caption{Sanity check metrics. $F$ is the feature activation frequency, $H$ is the entropy, and $A_s$ (respectively $A_t$) is the activation rate on the different squares (respectively trajectories). $\mathcal{P}$ is a linear probe trained to differentiate optimality, with F-score ($F_1$), precision $P$ and recall ($R$).
    }
    \label{tab:sanity_metrics}
\end{table}
We report more dead (frequency $F<0.1\%$) $c$-features, i.e. an over-specification of the $c$-features, and more overactive (frequency $F>10\%$) $d$-features, i.e. over-generalisation of $d$-features.
We see that the entropy $H(A_s)$, the entropy of activation distribution over the square, and respectively $H(A_t)$, the entropy over the trajectories, is smaller for $c$-features, especially for trajectories. The $c$-features have overfitted certain trajectories, making them sort of look-up tables. 
% As a baseline, the maximum entropy achievable are respectively $\max H(A_s)=log(64)\approx4.16$ and $\max H(A_t)=log(500)\approx6.21$.
Finally, we train a linear classifier to find the difference between activations originating from optimal or suboptimal trajectories.
Notably, the probe $\mathcal{P}$ performances are better using $c$-features than $d$-features. 

\begin{figure}[H]
     \centering
     \begin{subfigure}[b]{0.49\textwidth}
         \centering
         \includegraphics[width=\textwidth]{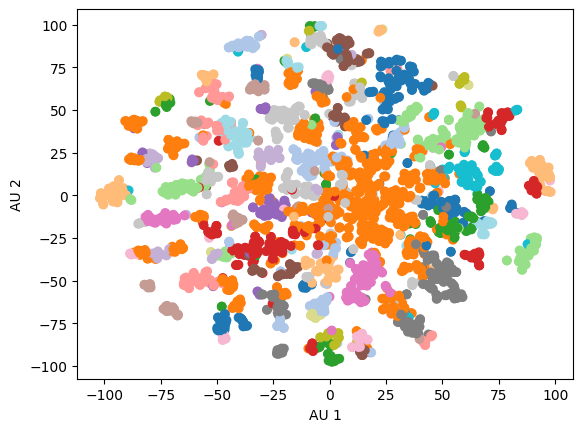}
         \caption{$c$-features clustering}
     \end{subfigure}
     \begin{subfigure}[b]{0.49\textwidth}
         \centering
         \includegraphics[width=\textwidth]{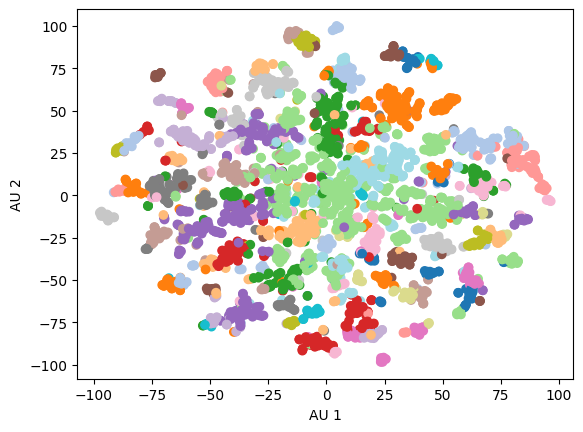}
         \caption{$d$-features clustering}
     \end{subfigure}
     \caption{Agglomerative clustering of the \texttt{test} samples after an NMF followed by a t-SNE for the visualisation \citep{Maaten2008VisualizingDU,scikit-learn}. We present the first 100 clusters, and colours are repeated. Each colour represents 5 different clusters, and the colours are independent of (a) and (b).
     While the structures are similar (due to the t-SNE projection), the labels are uncorrelated, suggesting a difference in representations for the $c$-features and $d$-features.
     }
     \label{fig:sanity_cls}
\end{figure}

\paragraph{Correlation of features}
To further compare the $c$-features and $d$-features, we clustered the samples using either of them. The visualisation, figure \ref{fig:sanity_cls}, looks alike for both, but the attribution of classes is uncorrelated, with a maximum person coefficient per cluster pair averaging over $0.1$.

To categorise the two clusterisation approaches, we explored the cluster specificity with respect to the square, state optimality, and trajectory. For that, we computed the respective entropy $H_s$, $H_o$, and $H_t$ for each cluster, reported in table \ref{tab:sanity_metrics_cls}.
We found no clear distinction between the two clusterisations. This informs us that both sets of features contain overspecific features that should be removed, as reported in appendix \ref{sec:unwanted_features}, but overall, they can be used in combination.

\begin{table}[H]
    \centering
    \begin{tabular}{l|c|c|c|}
    Metric &  $H_s$&$H_o$ &$H_t$
    \\
    \hline
    $c$-features &
    $2.2\pm1.0$&
    $2.5\pm1.3$&
    $0.57\pm0.23$
    \\
    $d$-features &
    $2.53\pm0.92$&
    $2.9\pm1.1$&
    $0.62\pm0.17$
    \end{tabular}
    \caption{Entropy measures across the clusters of figure \ref{fig:sanity_cls} (mean and standard deviation). 
    }
    \label{tab:sanity_metrics_cls}
\end{table}

\subsection{Feature Interpretation}
\label{sec:feature_interp}

\begin{figure}[H]
     \centering
     \begin{subfigure}[b]{0.4\textwidth}
         \centering
         \includegraphics[width=\textwidth]{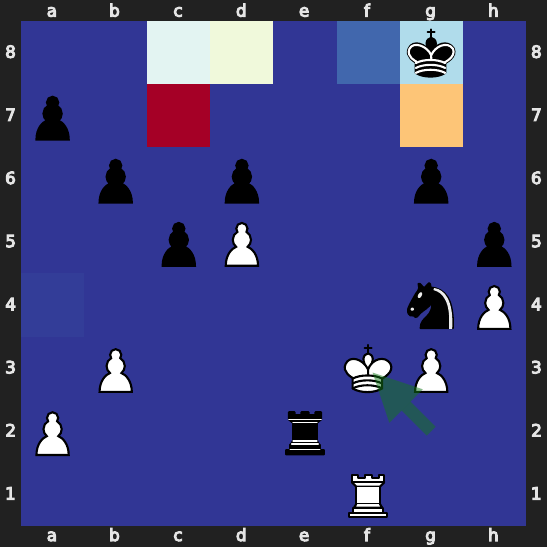}
         \caption{Safe place}
     \end{subfigure}
        \hspace{3em}
     \begin{subfigure}[b]{0.4\textwidth}
         \centering
         \includegraphics[width=\textwidth]{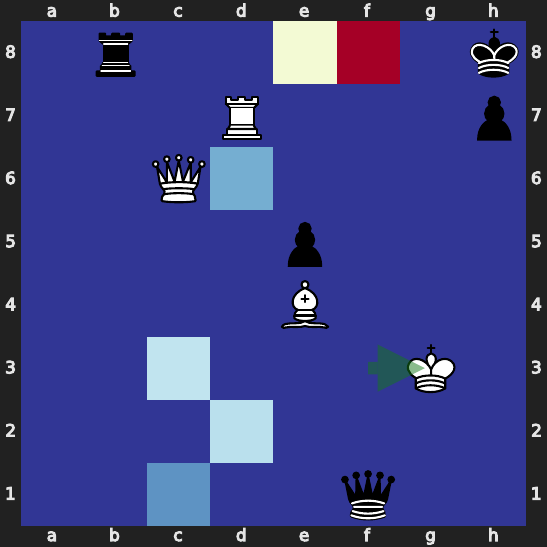}
         \caption{Protection}
     \end{subfigure}
     \caption{Illustration of a feature linked with the concept of protection. These samples were among the 16 samples that most activated the feature. On (a), the feature is activated on the king and a traditional safe place for the king. The path for the king to join the place is also activated. In (b), the black king is dangerously threatened, and a safe move might be to bring back the queen.}
     \label{fig:quali_1}
\end{figure}

\begin{figure}[H]
     \centering
     \begin{subfigure}[b]{0.4\textwidth}
         \centering
         \includegraphics[width=\textwidth]{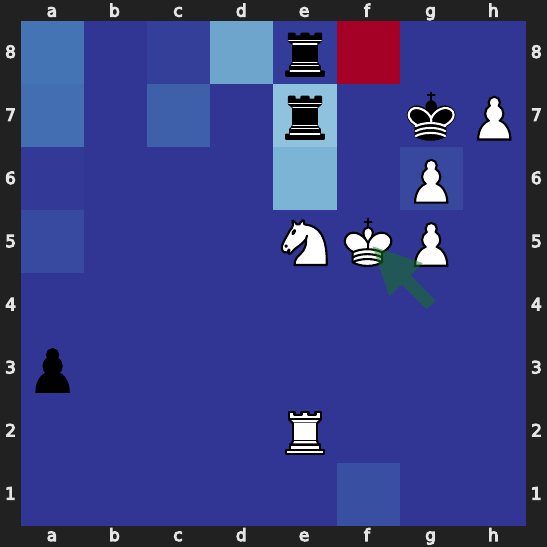}
         \caption{Rook threat 1}
     \end{subfigure}
     \hspace{3em}
     \begin{subfigure}[b]{0.4\textwidth}
         \centering
         \includegraphics[width=\textwidth]{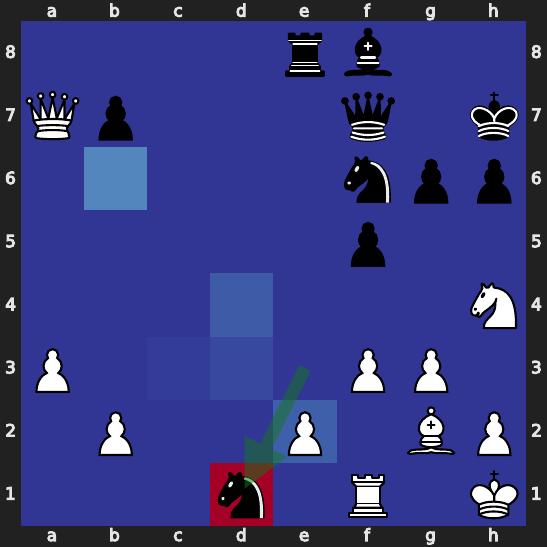}
         \caption{Rook threat 2}
     \end{subfigure}
     \caption{Illustration of a feature that seems to be linked with the concept of rook threat These samples were among the 16 samples that most activated the feature. The feature activates for both black and white. In (a), the black rook should move to the red square to check the king, while in (b), the white rook should take the knight.
     }
     \label{fig:quali_2}
\end{figure}

\paragraph{Qualitative Concept Analysis}
We cherry-picked features and the samples that maximally activated them to present qualitative analyses. The samples are selected by finding the maximally activating channels and computing the feature on their full board. We first present in the figure \ref{fig:quali_1} a feature that seemed to be linked to the pieces' safety.
And we then present a rook threat feature in figure \ref{fig:quali_2}.

\subsection{Dynamic Concept Clustering}
\label{sec:dyn_con_clustering}
\begin{figure}[H]
     \centering
     \begin{subfigure}[b]{0.49\textwidth}
         \centering
         \includegraphics[width=\textwidth]{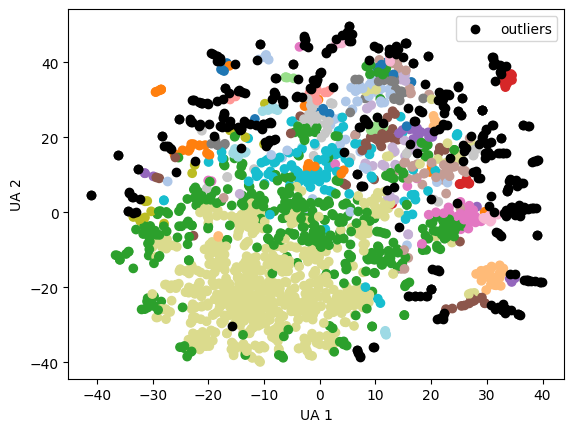}
         \caption{Clustered features}
     \end{subfigure}
     \begin{subfigure}[b]{0.49\textwidth}
         \centering
         \includegraphics[width=\textwidth]{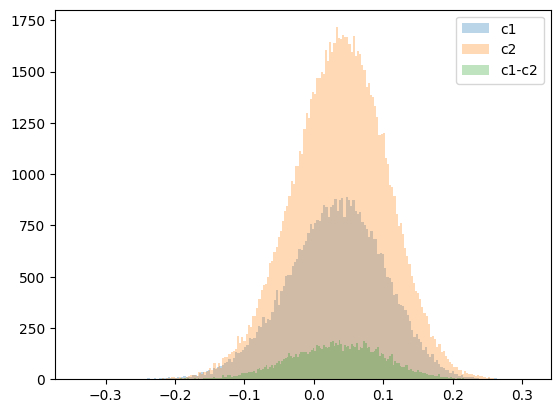}
         \caption{$W_d$ cosine similarities}
     \end{subfigure}
     \caption{(a) Clustering of the elicited features using an agglomerative clustering approach after an NMF followed by a t-SNE for the visualisation. (b) Cosine similarities of feature vectors originating from two significant clusters. There is no correlation
     between the intra and extra-cluster similarities.
     }
     \label{fig:conept_cls}
\end{figure}

\begin{figure}[H]
    \centering
    \includegraphics[width=0.6\textwidth]{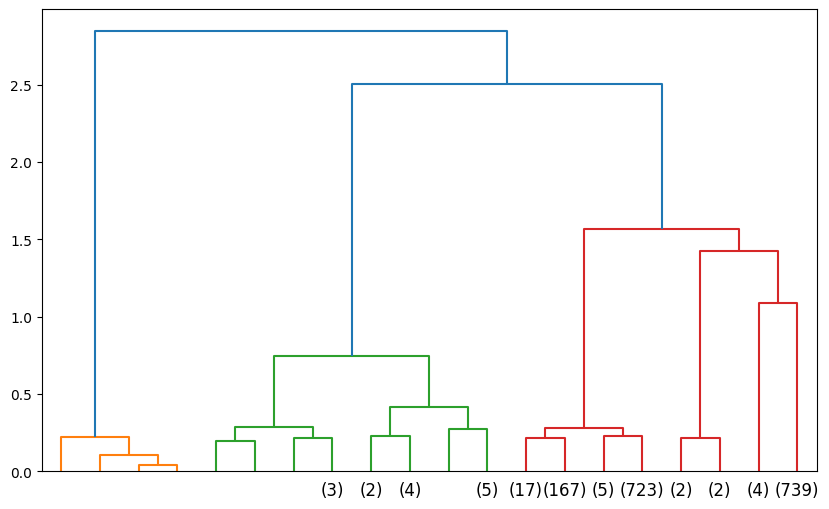}
     \caption{Dendrogram of the clustered features. The dendrogram can help visualise features and be leveraged to explore and interpret groups of features.}
     \label{fig:dend_cls}
\end{figure}

We present a way to explore features by grouping them. For that, we used an agglomerative clustering of features and reported the results in figure \ref{fig:conept_cls}. It seems here that a lot of features are outliers, but overall clusters appear. We found that the cluster can be found on the activation patterns of the feature, but it is not possible to use the feature vectors, i.e., the columns of $W_d$.
Finally, we report a dendrogram in figure \ref{fig:dend_cls}, i.e. an automated taxonomy of our elicited features. This analysis could be leveraged to adopt a more or less-grained view of the feature dictionary and thus explore it easily.
% This is especially important since a human in the loop still needs to decipher the meaning of the features.

\section{Discussion}

\subsection{Limitations}
\label{sec:lim} 

\paragraph{Having good SAEs}
SAE is still an active field of research, and there is an ongoing effort to find better training strategies and extract the most knowledge from them. It has also proven to be a challenge in this article. We present certain unwanted features in appendix \ref{sec:unwanted_features}.

\paragraph{Feature interpretation}
In order to interpret the features, human analysis cannot be totally replaced. We presented automated analyses in addition to our qualitative results, and we are excited about automated interpretability methods. Yet, having a human in the loop might be the only way not to defer to yet another black box, especially if it requires expert knowledge.

\paragraph{Contrastive interpretations}
Here, we didn't focus all our attention on finding contrastive interpretations, e.g. comparing the heatmap obtained on the root board and the trajectory board. Yet they might be more prominent, naturally emerging from our contrastive architecture. Thus, we should aim to interpret the features in a pair of root and trajectory visualisation.
In this respect features also show a blinking problem, i.e. features can have a different facet for white and black. Indeed, two similar boards will be encoded quite differently for white and black since the board is flipped for black. Because of this, we might need to pair black root boards with black trajectory boards.

\subsection{Future Work}

\paragraph{Concept sampling}
While we presented our sampling results in the appendix \ref{sec:technical_details}, our choices might have introduced inductive biases. It would be important to quantify the impact of different strategies for suboptimal sampling. For example, it is unclear to what extent the pairing strategy should take deeper trajectory boards and to what extent optimal and suboptimal trajectories can share a common state path.

\paragraph{Weak-to-strong generalisation}
We already mentioned that using a pair of latent activations is a more flexible interpretability method. But to go further, it is also possible to use the latent activation of smaller models to explain bigger models' strategies, as depicted by \cite{Burns2023WeaktoStrongGE}. While we only covered an introductory analysis, we think this track is highly promising and relevant to the safety of such models.

\paragraph{Different architectures}
A direct extension of this work would be to apply the same methodology to a model with the same architecture but a different number of layers. The scaling law could be compared across models w.r.t. the ELO and layer.
Furthermore, it would be interesting to use SAEs with a common feature dictionary and a specific encoder and decoder layer for each layer and checkpoint to compare feature transferability.

\section{Related Work}
\label{sec:related_work}

\paragraph{Discovering concepts in DNNs}
Linear probing is a simple idea where you train a linear model (probe) to predict a concept from the internals of the interpreted target model \citep{alain2018understanding}. The prediction performances are then attributed to the knowledge contained in the target model's latent representation rather than to the simple linear probe. 
In practice, a lasso formulation, i.e. $l_1$ penalty, has been a default choice as it encourages sparsity \citep{tibshirani1996regression}, and has been augmented as sparse probing for neuron attribution \citep{gurnee2023finding}.
Linear probing has also been derived with concept activation vectors \citep{kim2018interpretability}, which often require training a linear probe \citep{dreyer2023hope}.

\paragraph{Explaining chess models}
Chess has always been a good playground for AI, and explanability is no exception \citep{McGrath2022Acquisition}. Simplified versions of this game have even been created to make research easy \citep{hammersborg2023reinforcement,hammersborg2023information}.
It is even possible to explore planning, including tree search, through dynamical concepts \citep{schut2023bridging}.

\paragraph{Explainable tree search}
It is possible to make tree search explainable by default. By extracting a policy using a surrogate model \citep{Soemers2022ExtractingTL} or using a simpler heuristic model \citep{Soemers2019BiasingMW}.

\section{Conclusion}

This article explored multiple approaches to gaining knowledge from superhuman chess agents. We designed principles to try to elicit knowledge from the neural network's latent spaces. We successfully found interpretable features that were linked to the model plans. Furthermore, we proposed an automated feature taxonomy to help explore features, keeping a human in the loop. While presenting our key results, we also showed automated sanity checks. Finally, we presented the limitations and possible future directions to tackle them or to continue this project.

%%%%%%%%%%%%%%%%%%%%%%%%%%%%%%%%%%%%%%%%%%%%%%%%%%%%%%%%%%%%%%%%
%% Bibliography
%%%%%%%%%%%%%%%%%%%%%%%%%%%%%%%%%%%%%%%%%%%%%%%%%%%%%%%%%%%%%%%%
\bibliography{main}

\begin{thebibliography}{39}
\providecommand{\natexlab}[1]{#1}
\providecommand{\url}[1]{\texttt{#1}}
\expandafter\ifx\csname urlstyle\endcsname\relax
  \providecommand{\doi}[1]{doi: #1}\else
  \providecommand{\doi}{doi: \begingroup \urlstyle{rm}\Url}\fi

\bibitem[Alain \& Bengio(2018)Alain and Bengio]{alain2018understanding}
Guillaume Alain and Yoshua Bengio.
\newblock Understanding intermediate layers using linear classifier probes, 2018.

\bibitem[Bills et~al.(2023)Bills, Cammarata, Mossing, Tillman, Gao, Goh, Sutskever, Leike, Wu, and Saunders]{bills2023language}
Steven Bills, Nick Cammarata, Dan Mossing, Henk Tillman, Leo Gao, Gabriel Goh, Ilya Sutskever, Jan Leike, Jeff Wu, and William Saunders.
\newblock Language models can explain neurons in language models.
\newblock \url{https://openaipublic.blob.core.windows.net/neuron-explainer/paper/index.html}, 2023.

\bibitem[Bricken et~al.(2023)Bricken, Templeton, Batson, Chen, Jermyn, Conerly, Turner, Anil, Denison, Askell, Lasenby, Wu, Kravec, Schiefer, Maxwell, Joseph, Hatfield-Dodds, Tamkin, Nguyen, McLean, Burke, Hume, Carter, Henighan, and Olah]{bricken2023monosemanticity}
Trenton Bricken, Adly Templeton, Joshua Batson, Brian Chen, Adam Jermyn, Tom Conerly, Nick Turner, Cem Anil, Carson Denison, Amanda Askell, Robert Lasenby, Yifan Wu, Shauna Kravec, Nicholas Schiefer, Tim Maxwell, Nicholas Joseph, Zac Hatfield-Dodds, Alex Tamkin, Karina Nguyen, Brayden McLean, Josiah~E Burke, Tristan Hume, Shan Carter, Tom Henighan, and Christopher Olah.
\newblock Towards monosemanticity: Decomposing language models with dictionary learning.
\newblock \emph{Transformer Circuits Thread}, 2023.
\newblock https://transformer-circuits.pub/2023/monosemantic-features/index.html.

\bibitem[Burns et~al.(2022)Burns, Ye, Klein, and Steinhardt]{Burns2022DiscoveringLK}
Collin Burns, Haotian Ye, Dan Klein, and Jacob Steinhardt.
\newblock Discovering latent knowledge in language models without supervision.
\newblock \emph{ArXiv}, abs/2212.03827, 2022.
\newblock URL \url{https://api.semanticscholar.org/CorpusID:254366253}.

\bibitem[Burns et~al.(2023)Burns, Izmailov, Kirchner, Baker, Gao, Aschenbrenner, Chen, Ecoffet, Joglekar, Leike, Sutskever, Wu, and OpenAI]{Burns2023WeaktoStrongGE}
Collin Burns, Pavel Izmailov, Jan~Hendrik Kirchner, Bowen Baker, Leo Gao, Leopold Aschenbrenner, Yining Chen, Adrien Ecoffet, Manas Joglekar, Jan Leike, Ilya Sutskever, Jeff Wu, and OpenAI.
\newblock Weak-to-strong generalization: Eliciting strong capabilities with weak supervision.
\newblock \emph{ArXiv}, abs/2312.09390, 2023.
\newblock URL \url{https://api.semanticscholar.org/CorpusID:266312608}.

\bibitem[Campbell et~al.(2002)Campbell, Hoane, and hsiung Hsu]{campbell2002deep}
Murray Campbell, A.Joseph Hoane, and Feng hsiung Hsu.
\newblock Deep blue.
\newblock \emph{Artificial Intelligence}, 134\penalty0 (1):\penalty0 57--83, 2002.
\newblock ISSN 0004-3702.
\newblock \doi{https://doi.org/10.1016/S0004-3702(01)00129-1}.
\newblock URL \url{https://www.sciencedirect.com/science/article/pii/S0004370201001291}.

\bibitem[Chen et~al.(2020)Chen, Bei, and Rudin]{chen2020concept}
Zhi Chen, Yijie Bei, and Cynthia Rudin.
\newblock Concept whitening for interpretable image recognition.
\newblock \emph{Nature Machine Intelligence}, 2\penalty0 (12):\penalty0 772–782, December 2020.
\newblock ISSN 2522-5839.
\newblock \doi{10.1038/s42256-020-00265-z}.
\newblock URL \url{http://dx.doi.org/10.1038/s42256-020-00265-z}.

\bibitem[Conerly et~al.(2024)Conerly, Templeton, Bricken, Marcus, and Henighan]{Conerly2024Update}
Tom Conerly, Adly Templeton, Trenton Bricken, Jonathan Marcus, and Tom Henighan.
\newblock Update on how we train saes.
\newblock 2024.
\newblock URL \url{https://transformer-circuits.pub/2024/april-update/index.html}.

\bibitem[Conmy et~al.(2023)Conmy, Mavor-Parker, Lynch, Heimersheim, and Garriga-Alonso]{Conmy2023TowardsAC}
Arthur Conmy, Augustine~N. Mavor-Parker, Aengus Lynch, Stefan Heimersheim, and Adri{\`a} Garriga-Alonso.
\newblock Towards automated circuit discovery for mechanistic interpretability.
\newblock \emph{ArXiv}, abs/2304.14997, 2023.
\newblock URL \url{https://api.semanticscholar.org/CorpusID:258418244}.

\bibitem[Coulom(2006)]{Coulom2006EfficientSA}
R{\'e}mi Coulom.
\newblock Efficient selectivity and backup operators in monte-carlo tree search.
\newblock In \emph{Computers and Games}, 2006.
\newblock URL \url{https://api.semanticscholar.org/CorpusID:16724115}.

\bibitem[Cunningham et~al.(2023)Cunningham, Ewart, Riggs, Huben, and Sharkey]{Cunningham2023SparseAF}
Hoagy Cunningham, Aidan Ewart, Logan Riggs, Robert Huben, and Lee Sharkey.
\newblock Sparse autoencoders find highly interpretable features in language models.
\newblock \emph{ArXiv}, abs/2309.08600, 2023.
\newblock URL \url{https://api.semanticscholar.org/CorpusID:261934663}.

\bibitem[Dreyer et~al.(2023)Dreyer, Pahde, Anders, Samek, and Lapuschkin]{dreyer2023hope}
Maximilian Dreyer, Frederik Pahde, Christopher~J. Anders, Wojciech Samek, and Sebastian Lapuschkin.
\newblock From hope to safety: Unlearning biases of deep models via gradient penalization in latent space, 2023.

\bibitem[Gurnee et~al.(2023)Gurnee, Nanda, Pauly, Harvey, Troitskii, and Bertsimas]{gurnee2023finding}
Wes Gurnee, Neel Nanda, Matthew Pauly, Katherine Harvey, Dmitrii Troitskii, and Dimitris Bertsimas.
\newblock Finding neurons in a haystack: Case studies with sparse probing.
\newblock \emph{Transactions on Machine Learning Research}, 2023.
\newblock ISSN 2835-8856.
\newblock URL \url{https://openreview.net/forum?id=JYs1R9IMJr}.

\bibitem[Hammersborg \& Str{\"u}mke(2023{\natexlab{a}})Hammersborg and Str{\"u}mke]{hammersborg2023information}
Patrik Hammersborg and Inga Str{\"u}mke.
\newblock Information based explanation methods for deep learning agents--with applications on large open-source chess models.
\newblock \emph{arXiv preprint arXiv:2309.09702}, 2023{\natexlab{a}}.

\bibitem[Hammersborg \& Str{\"u}mke(2023{\natexlab{b}})Hammersborg and Str{\"u}mke]{hammersborg2023reinforcement}
Patrik Hammersborg and Inga Str{\"u}mke.
\newblock Reinforcement learning in an adaptable chess environment for detecting human-understandable concepts.
\newblock \emph{IFAC-PapersOnLine}, 56\penalty0 (2):\penalty0 9050--9055, 2023{\natexlab{b}}.

\bibitem[He et~al.(2016)He, Zhang, Ren, and Sun]{he2016deep}
Kaiming He, Xiangyu Zhang, Shaoqing Ren, and Jian Sun.
\newblock Deep residual learning for image recognition.
\newblock In \emph{2016 IEEE Conference on Computer Vision and Pattern Recognition (CVPR)}, pp.\  770--778, 2016.
\newblock \doi{10.1109/CVPR.2016.90}.

\bibitem[Hu et~al.(2019)Hu, Shen, Albanie, Sun, and Wu]{hu2019squeezeandexcitation}
Jie Hu, Li~Shen, Samuel Albanie, Gang Sun, and Enhua Wu.
\newblock Squeeze-and-excitation networks, 2019.

\bibitem[Kim et~al.(2017)Kim, Wattenberg, Gilmer, Cai, Wexler, Vi{\'e}gas, and Sayres]{Kim2017InterpretabilityBF}
Been Kim, Martin Wattenberg, Justin Gilmer, Carrie~J. Cai, James Wexler, Fernanda~B. Vi{\'e}gas, and Rory Sayres.
\newblock Interpretability beyond feature attribution: Quantitative testing with concept activation vectors (tcav).
\newblock In \emph{International Conference on Machine Learning}, 2017.
\newblock URL \url{https://api.semanticscholar.org/CorpusID:51737170}.

\bibitem[Kim et~al.(2018)Kim, Wattenberg, Gilmer, Cai, Wexler, Viegas, and Sayres]{kim2018interpretability}
Been Kim, Martin Wattenberg, Justin Gilmer, Carrie Cai, James Wexler, Fernanda Viegas, and Rory Sayres.
\newblock Interpretability beyond feature attribution: Quantitative testing with concept activation vectors (tcav), 2018.

\bibitem[Kocsis \& Szepesv{\'a}ri(2006)Kocsis and Szepesv{\'a}ri]{kocsis2006bandit}
Levente Kocsis and Csaba Szepesv{\'a}ri.
\newblock Bandit based monte-carlo planning.
\newblock In Johannes F{\"u}rnkranz, Tobias Scheffer, and Myra Spiliopoulou (eds.), \emph{Machine Learning: ECML 2006}, pp.\  282--293, Berlin, Heidelberg, 2006. Springer Berlin Heidelberg.
\newblock ISBN 978-3-540-46056-5.

\bibitem[McGrath et~al.(2022)McGrath, Kapishnikov, Toma{\v{s} }ev, Pearce, Wattenberg, Hassabis, Kim, Paquet, and Kramnik]{McGrath2022Acquisition}
Thomas McGrath, Andrei Kapishnikov, Nenad Toma{\v{s} }ev, Adam Pearce, Martin Wattenberg, Demis Hassabis, Been Kim, Ulrich Paquet, and Vladimir Kramnik.
\newblock Acquisition of chess knowledge in {AlphaZero}.
\newblock \emph{Proceedings of the National Academy of Sciences}, 119\penalty0 (47), nov 2022.
\newblock \doi{10.1073/pnas.2206625119}.

\bibitem[Mikolov et~al.(2013)Mikolov, tau Yih, and Zweig]{Mikolov2013LinguisticRI}
Tomas Mikolov, Wen tau Yih, and Geoffrey Zweig.
\newblock Linguistic regularities in continuous space word representations.
\newblock In \emph{North American Chapter of the Association for Computational Linguistics}, 2013.
\newblock URL \url{https://api.semanticscholar.org/CorpusID:7478738}.

\bibitem[Nanda et~al.(2023)Nanda, Lee, and Wattenberg]{Nanda2023EmergentLR}
Neel Nanda, Andrew Lee, and Martin Wattenberg.
\newblock Emergent linear representations in world models of self-supervised sequence models.
\newblock \emph{ArXiv}, abs/2309.00941, 2023.
\newblock URL \url{https://api.semanticscholar.org/CorpusID:261530966}.

\bibitem[Nasu(2018)]{nasu2018nnue}
Yu~Nasu.
\newblock Nnue efficiently updatable neural-network based evaluation functions for computer shogi.
\newblock \emph{Ziosoft Computer Shogi Club}, 2018.

\bibitem[{Pascutto, Gian-Carlo and Linscott, Gary}(2019)]{pascutto2019leela}
{Pascutto, Gian-Carlo and Linscott, Gary}.
\newblock Leela chess zero, 2019.
\newblock URL \url{http://lczero.org/}.

\bibitem[Pedregosa et~al.(2011)Pedregosa, Varoquaux, Gramfort, Michel, Thirion, Grisel, Blondel, Prettenhofer, Weiss, Dubourg, Vanderplas, Passos, Cournapeau, Brucher, Perrot, and Duchesnay]{scikit-learn}
F.~Pedregosa, G.~Varoquaux, A.~Gramfort, V.~Michel, B.~Thirion, O.~Grisel, M.~Blondel, P.~Prettenhofer, R.~Weiss, V.~Dubourg, J.~Vanderplas, A.~Passos, D.~Cournapeau, M.~Brucher, M.~Perrot, and E.~Duchesnay.
\newblock Scikit-learn: Machine learning in {P}ython.
\newblock \emph{Journal of Machine Learning Research}, 12:\penalty0 2825--2830, 2011.

\bibitem[Radford et~al.(2015)Radford, Metz, and Chintala]{Radford2015UnsupervisedRL}
Alec Radford, Luke Metz, and Soumith Chintala.
\newblock Unsupervised representation learning with deep convolutional generative adversarial networks.
\newblock \emph{CoRR}, abs/1511.06434, 2015.
\newblock URL \url{https://api.semanticscholar.org/CorpusID:11758569}.

\bibitem[Rajamanoharan et~al.(2024)Rajamanoharan, Conmy, Smith, Lieberum, Varma, Kram'ar, Shah, and Nanda]{Rajamanoharan2024ImprovingDL}
Senthooran Rajamanoharan, Arthur Conmy, Lewis Smith, Tom Lieberum, Vikrant Varma, J'anos Kram'ar, Rohin Shah, and Neel Nanda.
\newblock Improving dictionary learning with gated sparse autoencoders.
\newblock 2024.
\newblock URL \url{https://api.semanticscholar.org/CorpusID:269362142}.

\bibitem[Rajendran et~al.(2024)Rajendran, Buchholz, Aragam, Sch{\"o}lkopf, and Ravikumar]{Rajendran2024LearningIC}
Goutham Rajendran, Simon Buchholz, Bryon Aragam, Bernhard Sch{\"o}lkopf, and Pradeep Ravikumar.
\newblock Learning interpretable concepts: Unifying causal representation learning and foundation models.
\newblock \emph{ArXiv}, abs/2402.09236, 2024.
\newblock URL \url{https://api.semanticscholar.org/CorpusID:267657802}.

\bibitem[Rosin(2011)]{Rosin2011MultiarmedBW}
Christopher~D. Rosin.
\newblock Multi-armed bandits with episode context.
\newblock \emph{Annals of Mathematics and Artificial Intelligence}, 61:\penalty0 203--230, 2011.
\newblock URL \url{https://api.semanticscholar.org/CorpusID:207081359}.

\bibitem[Schut et~al.(2023)Schut, Tomasev, McGrath, Hassabis, Paquet, and Kim]{schut2023bridging}
Lisa Schut, Nenad Tomasev, Tom McGrath, Demis Hassabis, Ulrich Paquet, and Been Kim.
\newblock Bridging the human-ai knowledge gap: Concept discovery and transfer in alphazero, 2023.

\bibitem[Shaham et~al.(2024)Shaham, Schwettmann, Wang, Rajaram, Hernandez, Andreas, and Torralba]{Shaham2024AMA}
Tamar~Rott Shaham, Sarah Schwettmann, Franklin Wang, Achyuta Rajaram, Evan Hernandez, Jacob Andreas, and Antonio Torralba.
\newblock A multimodal automated interpretability agent.
\newblock 2024.
\newblock URL \url{https://api.semanticscholar.org/CorpusID:269293025}.

\bibitem[Silver et~al.(2018)Silver, Hubert, Schrittwieser, Antonoglou, Lai, Guez, Lanctot, Sifre, Kumaran, Graepel, et~al.]{silver2018general}
David Silver, Thomas Hubert, Julian Schrittwieser, Ioannis Antonoglou, Matthew Lai, Arthur Guez, Marc Lanctot, Laurent Sifre, Dharshan Kumaran, Thore Graepel, et~al.
\newblock A general reinforcement learning algorithm that masters chess, shogi, and go through self-play.
\newblock \emph{Science}, 362\penalty0 (6419):\penalty0 1140--1144, 2018.

\bibitem[Soemers et~al.(2019)Soemers, Piette, and Browne]{Soemers2019BiasingMW}
Dennis J. N.~J. Soemers, {\'E}ric Piette, and Cameron Browne.
\newblock Biasing mcts with features for general games.
\newblock \emph{2019 IEEE Congress on Evolutionary Computation (CEC)}, pp.\  450--457, 2019.
\newblock URL \url{https://api.semanticscholar.org/CorpusID:84842738}.

\bibitem[Soemers et~al.(2022)Soemers, Samothrakis, Piette, and Stephenson]{Soemers2022ExtractingTL}
Dennis J. N.~J. Soemers, Spyridon Samothrakis, {\'E}ric Piette, and Matthew Stephenson.
\newblock Extracting tactics learned from self-play in general games.
\newblock \emph{Inf. Sci.}, 624:\penalty0 277--298, 2022.
\newblock URL \url{https://api.semanticscholar.org/CorpusID:255326863}.

\bibitem[Tibshirani(1996)]{tibshirani1996regression}
Robert Tibshirani.
\newblock Regression shrinkage and selection via the lasso.
\newblock \emph{Journal of the Royal Statistical Society Series B: Statistical Methodology}, 58\penalty0 (1):\penalty0 267--288, 1996.

\bibitem[Tigges et~al.(2023)Tigges, Hollinsworth, Geiger, and Nanda]{Tigges2023LinearRO}
Curt Tigges, Oskar~John Hollinsworth, Atticus Geiger, and Neel Nanda.
\newblock Linear representations of sentiment in large language models.
\newblock \emph{ArXiv}, abs/2310.15154, 2023.
\newblock URL \url{https://api.semanticscholar.org/CorpusID:264591569}.

\bibitem[Trager et~al.(2023)Trager, Perera, Zancato, Achille, Bhatia, and Soatto]{Trager2023LinearSO}
Matthew Trager, Pramuditha Perera, Luca Zancato, Alessandro Achille, Parminder Bhatia, and Stefan~0 Soatto.
\newblock Linear spaces of meanings: Compositional structures in vision-language models.
\newblock \emph{2023 IEEE/CVF International Conference on Computer Vision (ICCV)}, pp.\  15349--15358, 2023.
\newblock URL \url{https://api.semanticscholar.org/CorpusID:257766294}.

\bibitem[van~der Maaten \& Hinton(2008)van~der Maaten and Hinton]{Maaten2008VisualizingDU}
Laurens van~der Maaten and Geoffrey~E. Hinton.
\newblock Visualizing data using t-sne.
\newblock \emph{Journal of Machine Learning Research}, 9:\penalty0 2579--2605, 2008.
\newblock URL \url{https://api.semanticscholar.org/CorpusID:5855042}.

\end{thebibliography}
\bibliographystyle{rlc}

%%%%%%%%%%%%%%%%%%%%%%%%%%%%%%%%%%%%%%%%%%%%%%%%%%%%%%%%%%%%%%%%
%% Appendices
%%%%%%%%%%%%%%%%%%%%%%%%%%%%%%%%%%%%%%%%%%%%%%%%%%%%%%%%%%%%%%%%
\appendix
\section*{Appendix}
\section{Additional Chess Modelling Details}
\label{sec:addi_chess_modelling}

\paragraph{Board encoding}
The current position is encoded using planes, formally channels, equivalent to the colours in images, in a tensor of the shape $112\times8\times8$. The 112 planes can be first decomposed into two parts, the first 104 planes corresponding to the history planes (8 last boards) and 8 additional planes encoding the game metadata.
Each board of the history is encoded through 13 distinct planes, comprising two sets of 6 sparse planes each for the current\footnote{Note that the player is the same for all 8 boards of the history.} 
player's and the opponent's pieces, as illustrated in figure \ref{fig:background_a}.
The last 8 planes are always full planes and represent meta information like the castling rights, the current player's colour and the half-move clock value.

\paragraph{Move encoding}

The policy outputted by the network is a vector of size 1858. This number is obtained considering each starting position and counting all accessible ending positions using queen and knight moves. The different promotions should also be accounted for, with promotion to knight being the default in lc0. Note that as the corresponding moves are relative to the swapped board, promotion is only possible at rank 8. This table is hardcoded within the chess engine for programming efficiency and readability.

\paragraph{Tree-search} 

In practice, the $Q$-values $Q(s,a)$ are obtained through the value $V(s+a)$, and by adding the move-left-head utility $M_\theta(s+a)$ defined in equation \ref{eq:mlh}.
The value is simply computed using the network outputted probabilities and the defined reward $\mathcal{W}_\theta(s+a)\cdot R$. These engineering tricks make the network tuning flexible, e.g., to incentivise drawing or aiming for short games.

\begin{equation}
\label{eq:mlh}
    M(s+a)={\rm sign}(-V(s+a))
    \cdot\Pi_{m_{\rm max}}\left[m
    \cdot\left(\mathcal{M}_\theta(s+a)-\mathcal{M}_\theta(s)\right)
    \right]
    \cdot \chi \left[\overset{\sim}{V}(s+a)\right]
\end{equation}

With $\chi$ a second-degree polynomial function and $\overset{\sim}{V}$ the extra-value ratio defined as:

\begin{equation}
\label{eq:vmlh}
    \overset{\sim}{V}(s+a)={\rm ReLU}\left(\dfrac{|V(s+a)|-V_{\rm threshold}}{1-V_{\rm threshold}}\right)
\end{equation}

Here, the final bound used, equation \ref{eq:practical_upper_confidence_boundary}, doesn't rely on the visit could $N(s,a)$. It thus can be used with the raw output of the neural network to perform the sampling.

\begin{equation}
\label{eq:practical_upper_confidence_boundary}
    U(s,a)=\alpha V(s+a)+\beta M(s+a)+\gamma \mathcal{P}_\theta(s,a)
\end{equation}
\section{Technical Details}
\label{sec:technical_details}

\subsection{Dynamical Concepts Dataset}
\label{sec:dataset_details}

\paragraph{Chess boards dataset}
In order to train the SAEs, we created a base dataset\footnote{
Released at \href{https://huggingface.co/datasets/Xmaster6y/lczero-planning-tcec}{https://huggingface.co/datasets/Xmaster6y/lczero-planning-tcec}.
} of around 20k games from the TCEC archives. These games were then processed and transformed into 20M individual boards to form the board dataset\footnote{
Released at \href{https://huggingface.co/datasets/Xmaster6y/lczero-planning-boards}{https://huggingface.co/datasets/Xmaster6y/lczero-planning-boards}.
}. The first moves were filtered only to take position after the "book exits" and after at least 20 plys. For this preliminary study, we sampled trajectories from 200k random boards for the \texttt{train} split and 20k boards in the \texttt{test} split. The sampling of trajectories is further detailed below.

\paragraph{Concept sampling}
In order to choose the best strategy, i.e. the best hyperparameters of equation \ref{eq:practical_upper_confidence_boundary}, we run several matches between the different models and hyperparameters; the results are reported in table \ref{tab:hyper_turna}.
Using this strategy, we then constructed a trajectory dataset\footnote{
Released at \href{https://huggingface.co/datasets/Xmaster6y/lczero-planning-trajectories}{https://huggingface.co/datasets/Xmaster6y/lczero-planning-trajectories}.
} for each model. This dataset was then converted into an activation dataset\footnote{
Released at \href{https://huggingface.co/datasets/Xmaster6y/lczero-planning-activations}{https://huggingface.co/datasets/Xmaster6y/lczero-planning-activations}.
} to make the SAE training easy to configure. When sampling suboptimal trajectories, we used a normalised distribution without any optimal action.

\begin{table}[H]
    \centering
    \begin{tabular}{c|l|c|c|c|c|c}
         & &\multicolumn{5}{c}{Win rate vs $\mathcal{P}_\theta(s)$}  \\
        \hline 
        & Model & 1893 & 3051 & 4012 & 4238 & Average  \\
        \hline
        \multirow{8}{*}{\rotatebox[origin=c]{90}{Strategy}}& Raw $Q$-values: $\mathcal{W}_\theta(s+a)\cdot R$& $-0.18$ & $-0.48$ & $-0.73$ & $-0.78$ & $-0.55\pm0.24$ \\
&$U(s,a)$ ($\alpha=1$, $\beta=0$, $\gamma=0.25$)& $-0.17$ & $-0.45$ & $-0.65$ & $-0.63$ & $-0.48\pm0.19$ \\
&$U(s,a)$ ($\alpha=1$, $\beta=0$, $\gamma=0.5$)& $-0.10$ & $-0.35$ & $-0.67$ & $-0.48$ & $-0.40\pm0.21$ \\
&$U(s,a)$ ($\alpha=1$, $\beta=0$, $\gamma=1$)& $0.03$ & $0.03$ & $-0.13$ & $-0.15$ & $-0.05\pm0.09$ \\
&$U(s,a)$ ($\alpha=1$, $\beta=0.5$, $\gamma=0$)& $-0.18$ & $-0.57$ & $-0.73$ & $-0.68$ & $-0.54\pm0.22$ \\
&$U(s,a)$ ($\alpha=1$, $\beta=0.5$, $\gamma=0.1$)& $-0.20$ & $-0.43$ & $-0.72$ & $-0.68$ & $-0.51\pm0.21$ \\
&$U(s,a)$ ($\alpha=1$, $\beta=0.5$, $\gamma=0.25$)& $-0.07$ & $-0.37$ & $-0.67$ & $-0.65$ & $-0.44\pm0.25$ \\
&$U(s,a)$ ($\alpha=1$, $\beta=0.5$, $\gamma=0.5$)& $-0.12$ & $-0.33$ & $-0.55$ & $-0.43$ & $-0.36\pm0.16$
    \end{tabular}

    \caption{Hyperparameters tournament scores against the raw policy baseline. Only the combinations selected after an initial random search are reported. Here, the policy is better for almost all models and combinations.
    }
    \label{tab:hyper_turna}
\end{table}

\subsection{SAE Training}
\label{sec:training_details}

\begin{figure}[H]
     \centering
     \begin{subfigure}[b]{0.49\textwidth}
         \centering
         \includegraphics[width=\textwidth]{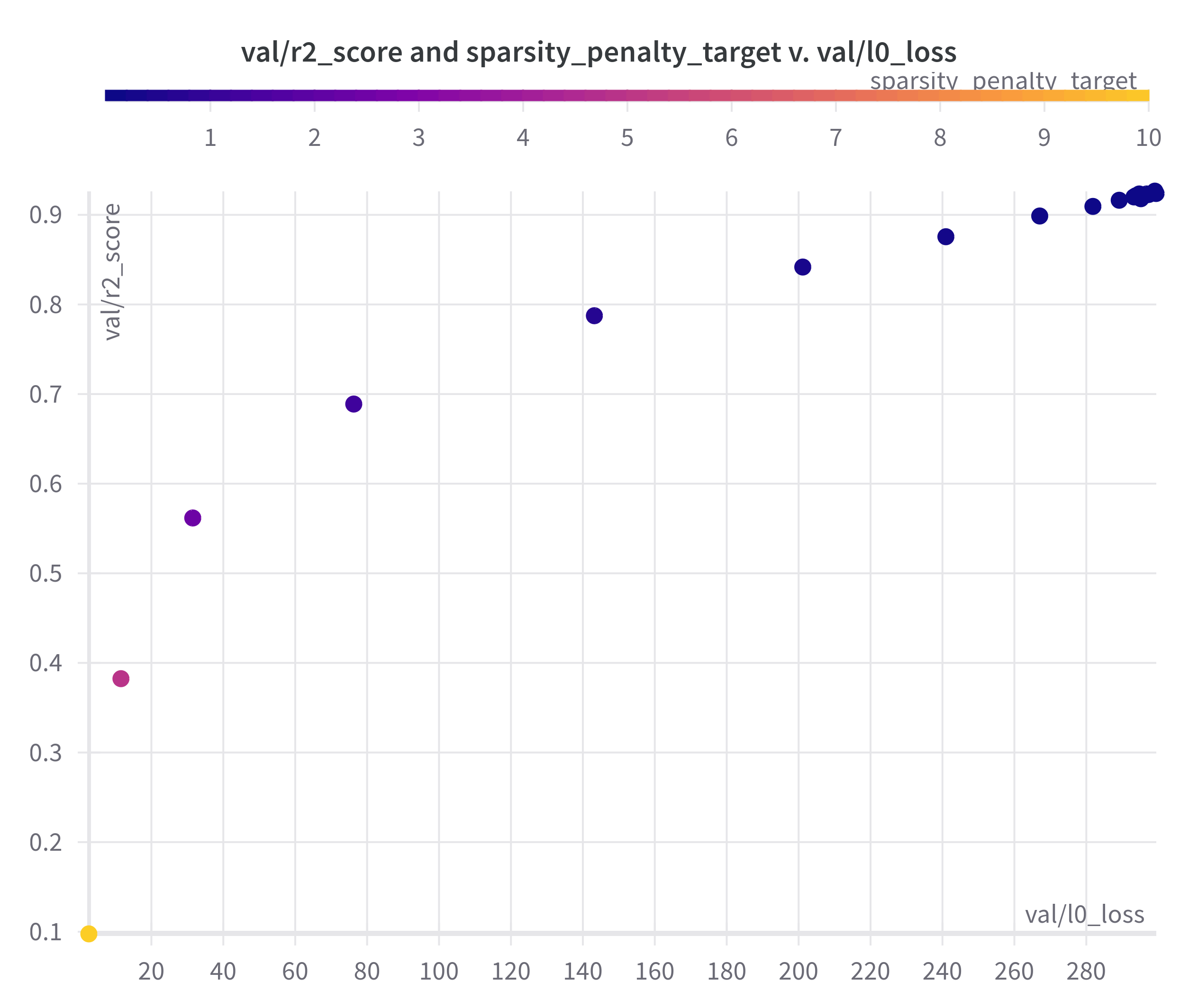}
         \caption{Trade-off sparsity/accuracy}
         \label{fig:trade-off}
     \end{subfigure}
     \begin{subfigure}[b]{0.49\textwidth}
         \centering
         \includegraphics[width=\textwidth]{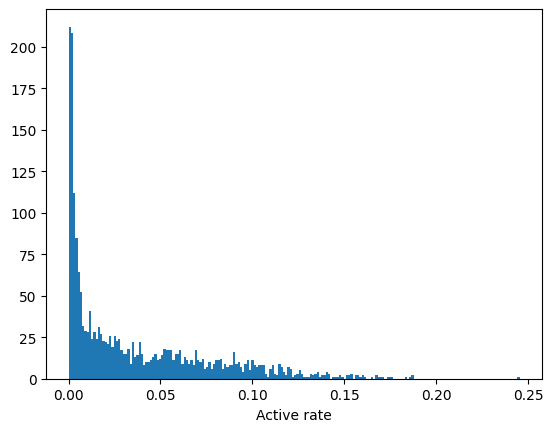}
         \caption{Feature activation histogram}
         \label{fig:feat_act}
     \end{subfigure}
     \caption{(a) Trade-off between the coefficient $R^2$ measuring the reconstruction accuracy vs the norm $\ell_0$ of the features, measuring the sparsity. The plot is obtained using a sweep of the coefficient $\lambda$ and shows a power law dependence. (b) The histogram of feature activation rate $F$. As already pointed out by previous works on SAE, a low-frequency cluster naturally emerges.}
     \label{fig:sae_training}
\end{figure}

\paragraph{Procedure}
We based our SAE training on recent work from like \cite{Rajamanoharan2024ImprovingDL} and take into account the monthly updates of Anthropic like \cite{Conerly2024Update}.
We will be reporting relevant metrics for our SAEs in the figure \ref{fig:sae_training}.
$\beta_1 = 0$ stabilised the training. We also use the modified loss, described in equation \ref{eq:sae_mod_loss}, in order to prevent arbitrary norm of dictionary columns that trick the $\ell_1$ norm. Indeed, without it, the features $f$ can get a low $\ell_1$ norm but not a low $\ell_0$ norm since even small features can reconstruct the activation $x$ if $W_d$ is unconstrained.

\begin{align}
    \label{eq:sae_mod_loss}
    \mathcal{L}_{\rm SAE}=\mathbb{E}_h\left[||h-\hat{h}||_2^2 +
    \lambda \sum_i|f_i|\cdot||{W_d}_i||_2
    \right]
\end{align}

We will release our trained assets\footnote{
Released at \href{https://huggingface.co/Xmaster6y/lczero-planning-saes}{https://huggingface.co/Xmaster6y/lczero-planning-saes}.
}. To make the SAE analysis easy, we also will release the feature activation dataset\footnote{
Released at \href{https://huggingface.co/datasets/Xmaster6y/lczero-planning-features}{https://huggingface.co/datasets/Xmaster6y/lczero-planning-features}.
} 
which will be then used in our interactive demonstration\footnote{
Accessible at \href{https://huggingface.co/spaces/Xmaster6y/lczero-planning-demo}{https://huggingface.co/spaces/Xmaster6y/lczero-planning-demo}.
}.
Hyperparameters are chosen to balance the trade-off between sparsity and reconstruction accuracy, as presented in the figure \ref{fig:trade-off}. We also monitor the activation of the feature, reported in figure \ref{fig:feat_act}, and as already discussed in the section \ref{sec:sanity_checks}.

\paragraph{Results}
When training SAEs, the first metrics to report, in addition to the losses, are the $\ell_0$ norm of features and the determination coefficient $R^2$ for the reconstruction. Indeed, we aim to jointly minimise the norm $\ell_0$ to get a sparse decomposition and maximise $R^2$ to ensure a correct reconstruction of the activations.
We showed in the table \ref{tab:losses} the different metrics obtained for the model used in this article. In particular, the trained SAE has, on average, 73 active features while trying to reconstruct activations of dimension 256, a reduction of around 71\%. But with respect to the dictionary, it represents only 3.5\% of active features. 

\begin{table}[H]
    \centering
    \begin{tabular}{l|c|c|c|c|c|c|c|}
    Losses &  MSE & Sparsity & $\mathcal{L}_{\rm contrast}$ & $\ell_0$ & $R^2$ 
    \\
    \hline
    \texttt{train} & 21.7 & 26.7 & 10.7 & 73.3 & 0.81 \\
    \texttt{validation} & 21.8 & 26.8 & 10.7 & 73.4 & 0.81 
    \end{tabular}
    \caption{Losses and metrics obtained for the model used in this article for the sets \texttt{train} and \texttt{validation}. MSE refers to the mean squared error, e.g. the reconstruction loss $\mathbb{E}_h\left[||h-\hat{h}||_2^2\right]$, and similarly Sparsity refers to $||f||_1$. $\ell_0$ and $R^2$ are metrics that were optimised using the \texttt{validation} set. $\ell_0$ measures the feature sparsity and $R^2$ the activation reconstruction (1 is the best). As $\ell_0$ is a count, it can be understood knowing that the activation dimension is 256 and the dictionary dimension is 2048.
    }
    \label{tab:losses}
\end{table}

\begin{figure}[H]
     \begin{subfigure}[b]{0.49\textwidth}
         \centering
         \includegraphics[width=\textwidth]{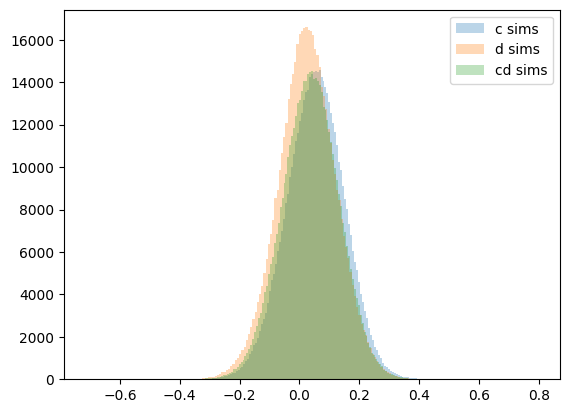}
         \caption{CSAE}
     \end{subfigure}
     \hfill
     \begin{subfigure}[b]{0.49\textwidth}
         \centering
         \includegraphics[width=\textwidth]{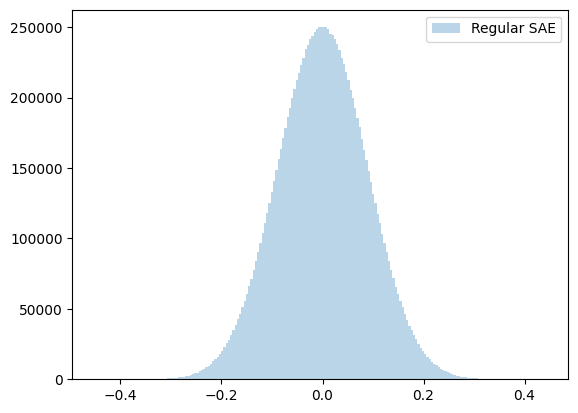}
         \caption{SAE}
     \end{subfigure}
     \caption{Histogram of the cosine similarities of the dictionary vectors. (a) is 
     reported for our CSAE and (b) for a regular SAE. We find that we conserve the 
     independence of the learned directions.
     }
     \label{fig:sims}
\end{figure}
\section{Concepts in Different Models and Layers}
\label{sec:accross_models_layers}

\begin{figure}[H]
     \centering
     \includegraphics[width=0.5\textwidth]{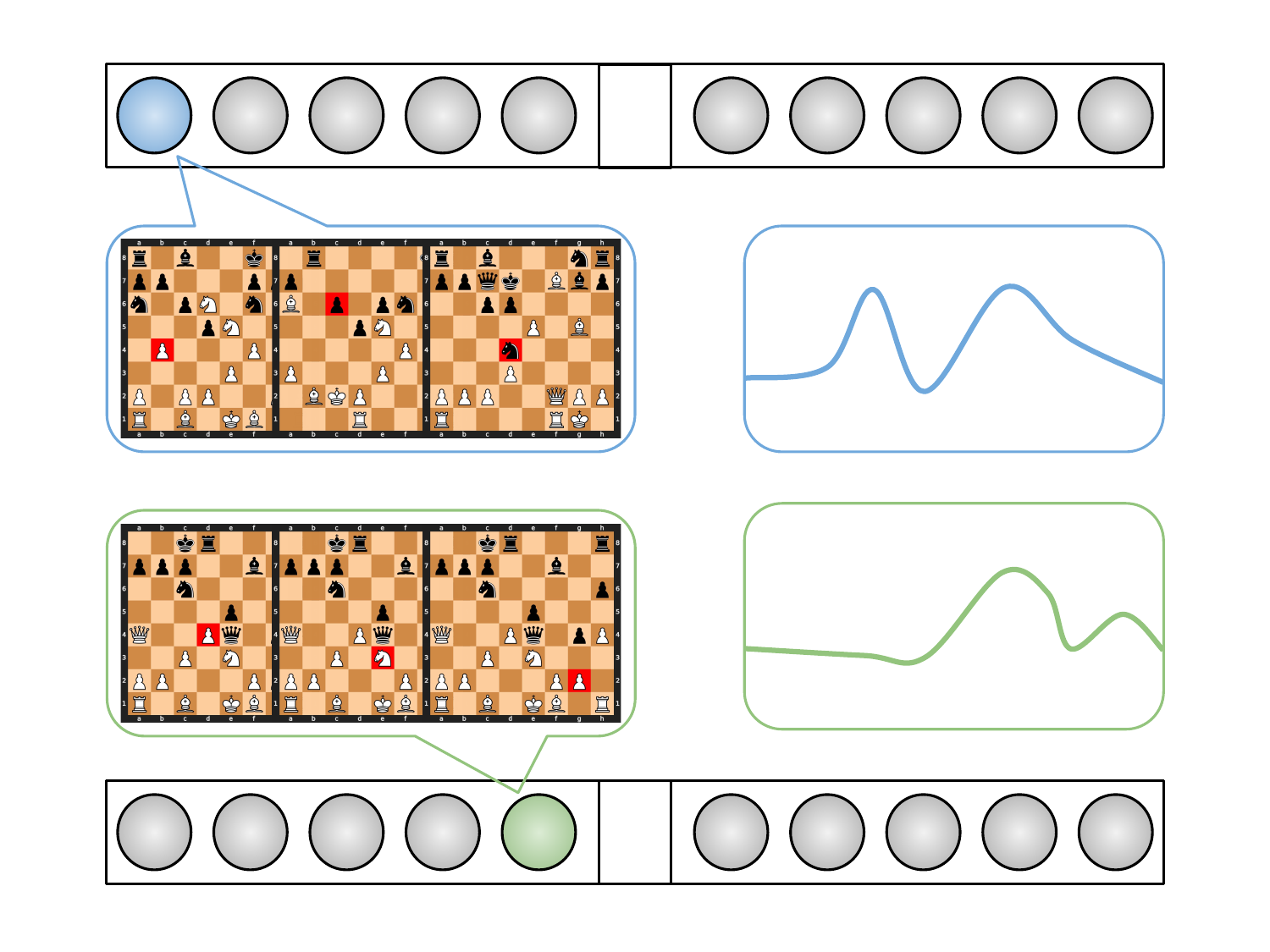}
     \caption{In order to compare a pair of features, the first indicator is the correlation of the feature activation (right). It is also possible to count common samples retrieved using activation maximisation (left).}
     \label{fig:act_max_comp}
\end{figure}

\paragraph{Comparing features by pair}
It is important to investigate the correlation between features, which is a simple proxy to understand basic interactions between features. This analysis can be run for the $c$-features and the $d$-features, which is illustrated in figure \ref{fig:act_max_comp}. We first present a sanity check on the $c$-features in section \ref{sec:sanity_checks} and expand $d$-features categorisation in \ref{sec:dyn_con_clustering}.
This method is especially relevant when dealing with different latent spaces, e.g. from different models or layers.  In the following paragraph, we present a small investigation of the correlation between features from different layers and at different training stages.

\paragraph{Probing across different latent spaces}
In order to investigate universal concepts shared across models or layers we need to probe different latent spaces. A quick analysis of these latent spaces yields that they differ, at least in barycentre, amplitude, and principal components. We thus only investigate the correlation between features and leave the design of universal SAEs decomposing multiple latent spaces simultaneously for future work.
Similarly to \cite{bricken2023monosemanticity}, to analyse features of different SAEs, we used the correlation of the activations to which we add the correlation between the most activated sample, i.e. using data-based activation maximisation \cite{chen2020concept}.

\begin{figure}[H]
     \begin{subfigure}[b]{0.49\textwidth}
         \centering
         \includegraphics[width=\textwidth]{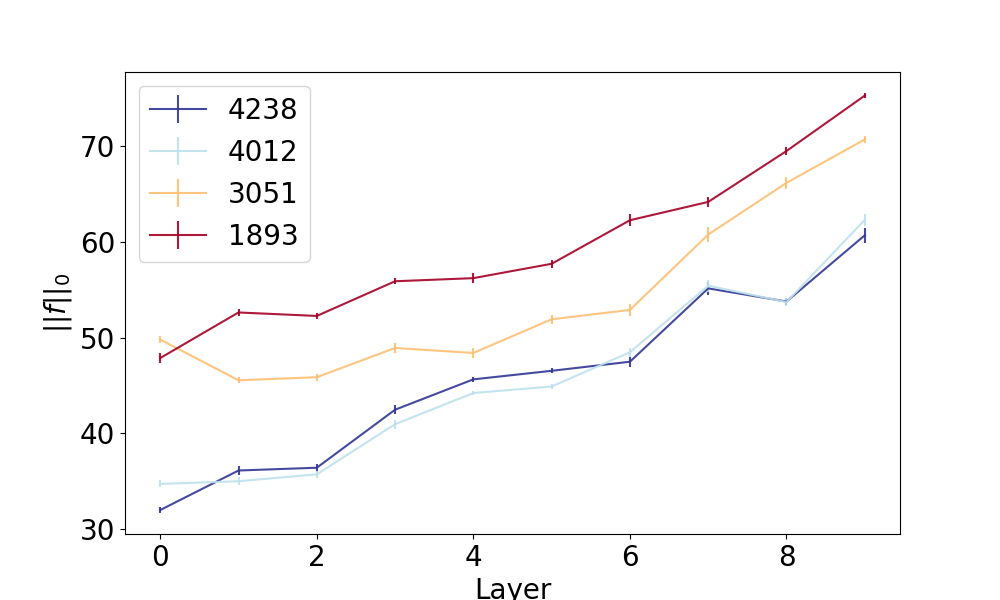}
         \caption{$||f||_0$ across layers for different models named after their ELO.}
     \end{subfigure}
     \hfill
     \begin{subfigure}[b]{0.49\textwidth}
         \centering
         \includegraphics[width=\textwidth]{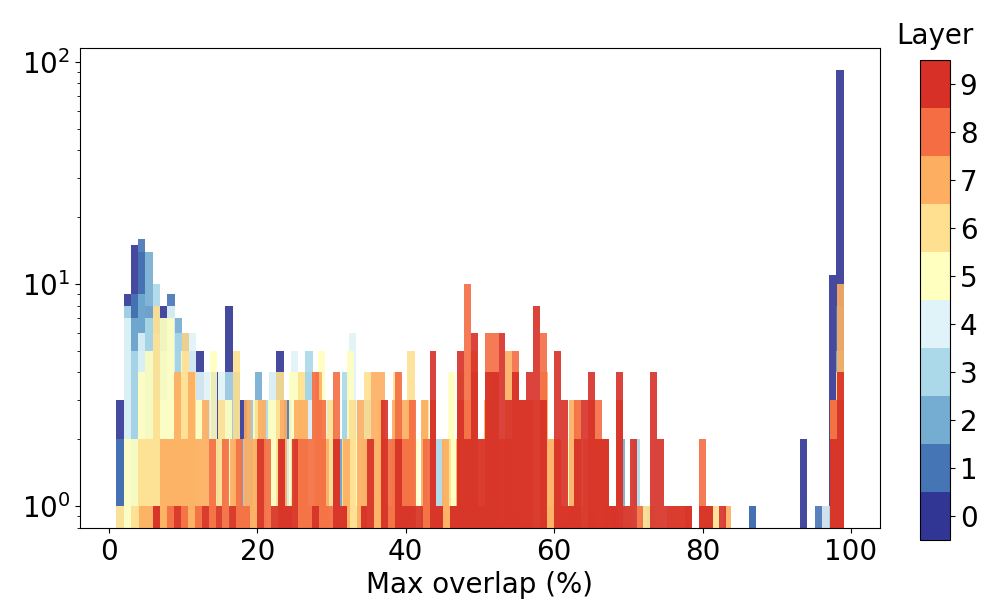}
         \caption{Overlap for ELO-4012 with the last layer of ELO-4238.}
     \end{subfigure}
     \caption{Feature analysis of the agents' latent spaces, summarising scaling properties. The SAEs trained for this figure are regular ones (without the contrastive framing). (a) represents the evolution of $\ell_0$ on different models and at different layers. There seems to be a general trend of information densification through layers but more condensed in better models. (b) represent the correlation between features of different layers. While the gradual correlations is expected to correlate with layers, the peak at 100\% could indicate over-active features or universal ones.
     }
     \label{fig:results}
\end{figure}

\paragraph{Feature comparison}
The study was on a 10-layer model across 4 checkpoints named after their ELO, i-e, their chess performance level; the results are shown in figure \ref{fig:results}. While conclusions must be drawn with care, Figure \ref{fig:results}(a) seems to show a scaling law of feature density or storage across layers and training. Later latent spaces are denser, surely due to refined and more complex information, but the training compresses the latent spaces, possibly using sharper features. Figure \ref{fig:results}(b) represents the correlation between maximum activated samples between the last layer of ELO-4238 and the layers of ELO-4012 and indicates that earlier layers wield more universal features.

\section{Unwanted Features}
\label{sec:unwanted_features}
We show two kinds of unwanted features that are present in our trained SAE.

\paragraph{Square specific features}Features that are specific to a given square. They act as over-generic features.

\begin{figure}[H]
     \centering
     \begin{subfigure}[b]{0.4\textwidth}
         \centering
         \includegraphics[width=\textwidth]{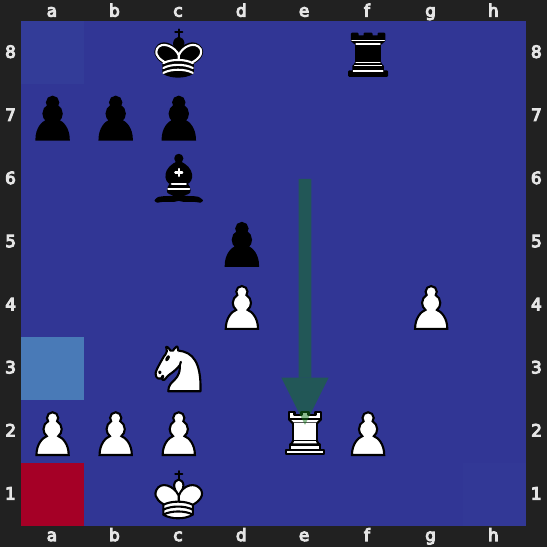}
         \caption{White facet}
     \end{subfigure}
        \hspace{3em}
     \begin{subfigure}[b]{0.4\textwidth}
         \centering
         \includegraphics[width=\textwidth]{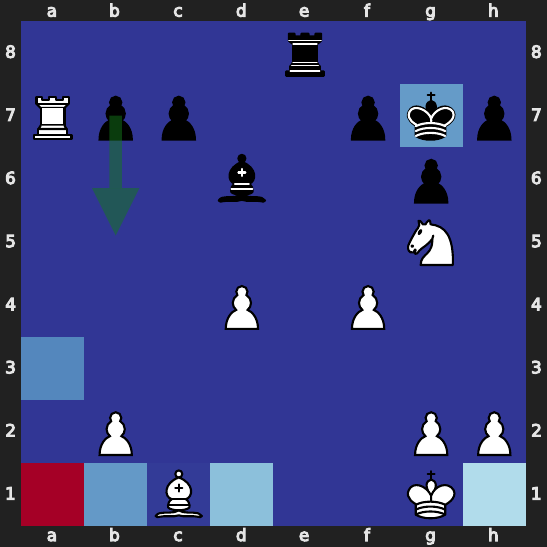}
         \caption{Black facet}
     \end{subfigure}
     \caption{Illustration of a feature that is linked to the lower left square (a1). (a) was among the 16 samples that most activated the feature, and (b) was chosen arbitrarily. The feature is sometimes dead or differently activated but mostly activates on a1. It also happens to activate on a8 relatively when the heatmap is when the heatmap is flipped according to the model's view.}
\end{figure}

\paragraph{Trajectory specific features} Features that are specific to a given trajectory. They act as lookup tables.

\begin{figure}[H]
     \centering
     \begin{subfigure}[b]{0.4\textwidth}
         \centering
         \includegraphics[width=\textwidth]{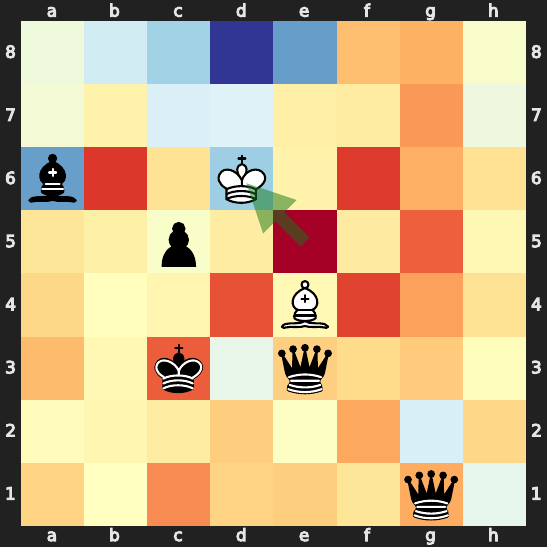}
         \caption{Specific trajectory state}
     \end{subfigure}
        \hspace{3em}
     \begin{subfigure}[b]{0.4\textwidth}
         \centering
         \includegraphics[width=\textwidth]{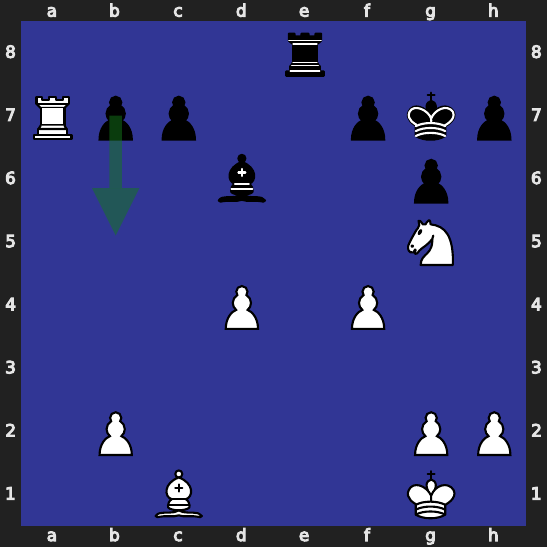}
         \caption{Protected square}
     \end{subfigure}
     \caption{Illustration of a feature that is linked to a particular trajectory. (a) was among the 16 samples that most activated the feature, and (b) was chosen arbitrarily. On (a), the feature is activated on almost every square, but on (b), it is dead.}
\end{figure}

\end{document}